\PassOptionsToPackage{table}{xcolor}
\documentclass{article} 
\usepackage{iclr2027_conference,times}
\usepackage[T1]{fontenc}

\usepackage{amsmath,amsfonts,bm}

\def\eqref#1{equation~\ref{#1}}

\def\1{\bm{1}}

\DeclareMathAlphabet{\mathsfit}{\encodingdefault}{\sfdefault}{m}{sl}
\SetMathAlphabet{\mathsfit}{bold}{\encodingdefault}{\sfdefault}{bx}{n}

\usepackage{url}
\usepackage{amsmath}
\usepackage{amssymb}
\usepackage{booktabs}
\usepackage{multirow}
\usepackage{graphicx}
\usepackage{xcolor}
\usepackage{makecell}
\usepackage{array}
\usepackage{tabularx}
\usepackage{longtable}
\usepackage{threeparttable}
\usepackage{microtype}
\usepackage{placeins}
\usepackage{float}
\usepackage{enumitem}
\usepackage{balance}
\usepackage[hidelinks]{hyperref}

\definecolor{asmblue}{HTML}{2F66E8}
\definecolor{asmlight}{HTML}{EEF4FF}
\definecolor{asmgroup}{HTML}{F4F6F8}
\definecolor{asmdark}{HTML}{15223A}
\definecolor{asmgray}{HTML}{5D6675}
\newcommand{\tabtitle}[1]{\textbf{#1}}
\newcommand{\tabhead}[1]{\textbf{#1}}
\newcommand{\tablegroup}[2]{%
  \rowcolor{asmgroup}\multicolumn{#1}{@{}l}{\textcolor{asmdark}{\bfseries #2}}\\}
\newcommand{\best}[1]{\textbf{#1}}

\newcommand{\ours}{\textsc{PReM}}
\newcommand{\gain}[1]{\textcolor{asmblue}{\bfseries$\uparrow$#1}}
\newcommand{\dropval}[1]{\textcolor{asmgray}{$\downarrow$#1}}

\renewcommand{\arraystretch}{1.08}


\title{PReM: Prefix-Steered Recurrent Memory \\ for Long-Video Understanding}

\author{%
Siru Zhong$^{1,2}$,
Qiongyan Wang$^{1}$,
Xiaohui Lv$^{2}$,
Yuzheng Zhuang$^{2}$,\\
\bfseries Shuai Tao$^{2,*}$,
Wulong Liu$^{2}$,
Haohuan Fu$^{3}$,
Yuxuan Liang$^{1,*}$\\
$^{1}$The Hong Kong University of Science and Technology (Guangzhou)\\
$^{2}$Beta Infinity \quad
$^{3}$Tsinghua University
}
\vspace{-1em}

\iclrfinalcopy 
\begin{document}
\raggedbottom

\maketitle

\begin{abstract}
Long-video understanding must capture transient visual evidence under strict token budgets, yet existing methods compress frames, append memory tokens, or alter internal key-value (KV) caches. We introduce \textbf{\underline{P}refix-Steered \underline{Re}current \underline{M}emory} (\textbf{\ours{}}), a memory-token-free framework for frozen vision-language models (VLMs). \ours{} separates video ingestion from query answering: a recurrent writer distills visual streams into a compact ${\approx}256$\,KiB multi-slot associative state, while a question-conditioned readout adds memory-derived key/value (K/V) steering modulations to existing non-visual prompt prefixes during prefill. This enables write-once, query-many inference without extra prompt tokens or decoding recurrence. Across six long-video benchmarks in offline and streaming end-of-stream settings, \ours{} consistently outperforms frozen baselines at every evaluated visual budget. Under a constrained budget of 16 frames, \ours{} improves macro-average accuracy by 3.06\% on Qwen2.5-VL-3B, with gains of 11.0\% on action antonym identification and 9.9\% on localized needle retrieval. These gains require tuning ${\approx}0.24\%$ of backbone parameters at 0.03\,GiB of peak GPU memory overhead. Code is available at \url{https://github.com/siruzhong/PReM}.
\end{abstract}
\section{Introduction}
\label{sec:intro}

Long-video understanding requires multimodal models to reason over visual evidence scattered across extended temporal horizons \citep{wu2024longvideobench,fu2025videomme,mangalam2023egoschema}. In practical domains such as continuous video streams, egocentric perception \citep{grauman2022ego4d}, instructional procedures \citep{tang2019coin}, and robotic manipulation and navigation \citep{brohan2023rt2,anwar2024remembr}, decisive evidence is often brief and surrounded by uninformative content. Effective reasoning thus demands both comprehensive temporal coverage and high sensitivity to subtle cues. However, modern vision-language models (VLMs) are constrained by strict visual token budgets due to quadratic attention complexity and GPU memory limits.

\begin{figure}[t]
  \centering
  \vspace{-1em}
  \includegraphics[width=\linewidth]{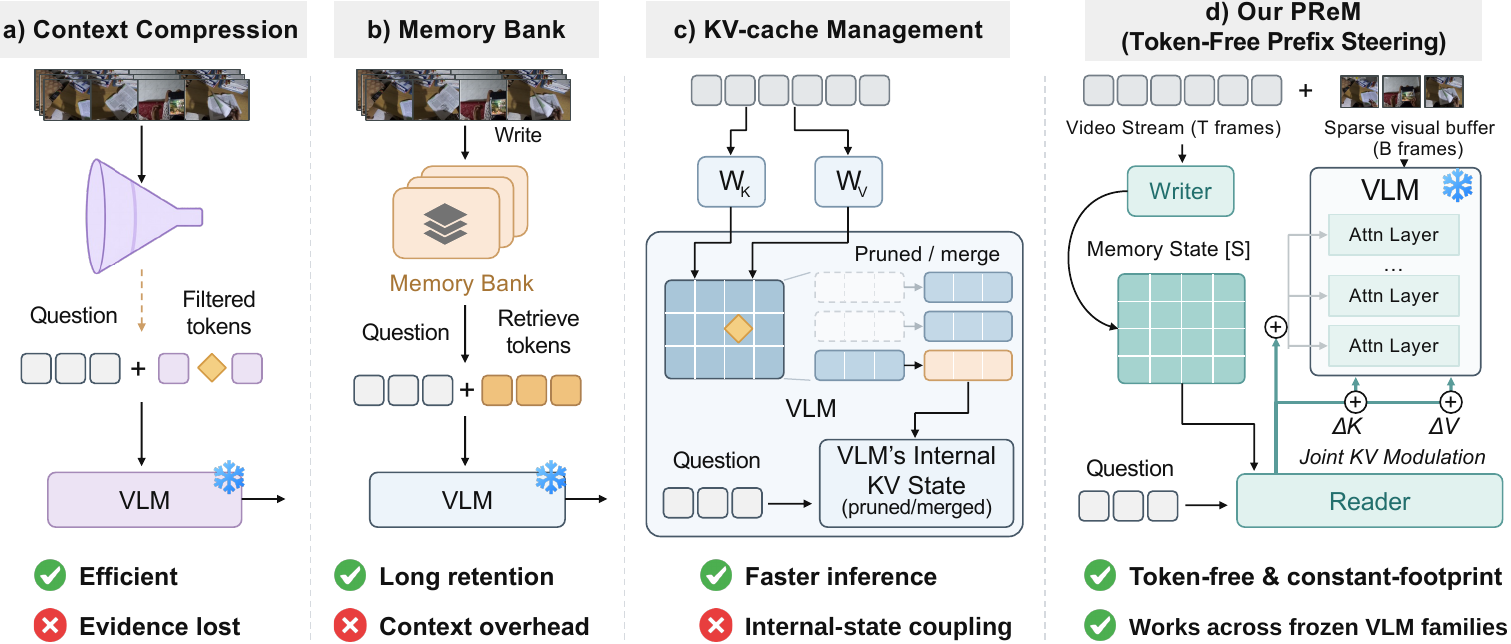}
  \caption{Comparison of long-video VLM paradigms. \textbf{(a)} Compression reduces visual context. \textbf{(b)} Memory banks consume prompt capacity. \textbf{(c)} KV management operates on internal attention states. \textbf{(d)} \ours{} steers the backbone via prefix modulations $(\Delta K, \Delta V)$ without sequence tokens.}
  \label{fig:interfaces}
  \vspace{-1em}
\end{figure}

This tension creates an inevitable trade-off between temporal breadth and sequence cost. Aggressive frame downsampling risks discarding transient events entirely, while admitting more visual tokens incurs steep computational overhead and dilutes attention, frequently inducing ``lost-in-the-middle'' failures \citep{liu2024lostmiddle}. As illustrated in Figure~\ref{fig:interfaces}, existing approaches navigate this trade-off through three main paradigms: \textbf{(1)~Visual context compression} reduces spatiotemporal redundancy via pooling, frame selection, or multi-rate sampling \citep{cheng2024videollama2,shen2025longvu,li2025videochatflash,xu2024slowfastllava}, yet information discarded during compression cannot be recovered for subsequent reasoning; \textbf{(2)~Video memory banks} maintain episodic feature pools or explicit token banks \citep{he2024malmm,song2024moviechat,wang2025videollamb,zhang2025flashvstream,diko2025rewind,liu2026beyond}, but their token-based readouts directly occupy prompt sequence capacity; and \textbf{(3)~KV-cache management} dynamically prunes, merges, or evicts intermediate attention caches \citep{qin2025videoxl2,kim2025infinipotv,di2025rekv,ning2025livevlm,he2025streammem,chen2026streamttt}, which avoids prompt token overhead but couples directly to internal attention history.

Consequently, historical visual evidence in existing paradigms either competes for prompt sequence budget or couples directly to internal attention caches. To address this challenge, we propose \textbf{Prefix-Steered Recurrent Memory (\ours{})}, a dual-pathway framework that separates video ingestion from query answering. During a one-time ingestion pass, a query-agnostic recurrent writer distills incoming visual features into a compact, multi-slot associative state (${\approx}256$\,KiB) using selective gating to filter redundancy and protect salient cues. At query time, a question-conditioned readout retrieves memory content and injects it as additive key/value (K/V) steering modulations into non-visual prompt prefixes exclusively during prefill. This design steers the frozen VLM's attention over its visual buffer without appending auxiliary prompt tokens or altering historical visual caches, supporting write-once, query-many inference.

Our main contributions are summarized as follows:
\begin{itemize}[leftmargin=*]
  \item \textbf{Decoupled Memory Architecture.} We introduce a memory-token-free framework for frozen VLMs that separates continuous video ingestion from query answering, retaining historical visual evidence in an external, constant-footprint recurrent state without consuming prompt token capacity.
  \item \textbf{Gated Recurrence \& Prefill-Only Steering.} We design a multi-factor gated write update to selectively filter temporal redundancy, coupled with a question-conditioned readout that steers prefix K/V representations during prefill, eliminating recurrent updates during generation.
  \item \textbf{Comprehensive Empirical Validation.} Across six long-video benchmarks and three VLM families, \ours{} improves macro-average accuracy at every evaluated budget in both offline and streaming end-of-stream settings. Under a constrained budget of 16 frames, it achieves a 3.06\% macro-average gain on Qwen2.5-VL-3B (with task-specific improvements up to 11.0\%) while tuning ${\approx}0.24\%$ parameters at 0.03\,GiB peak GPU memory overhead.
\end{itemize}

\section{Related Work}
\label{sec:related}

\noindent\textbf{Visual context compression.}
Visual context compression alleviates quadratic attention costs by pruning tokens along different dimensions. VideoLLaMA~2 \citep{cheng2024videollama2} aggregates spatiotemporal features via convolutional connectors, while LLaMA-VID \citep{li2023llamavid} distills each frame into dual context-content tokens. To balance spatial detail and motion, SlowFast-LLaVA \citep{xu2024slowfastllava} adopts asymmetric frame-rate pathways. Complementarily, LongVU \citep{shen2025longvu} and VideoChat-Flash \citep{li2025videochatflash} employ text-guided frame selection and hierarchical token compression, respectively, whereas LongVILA \citep{chen2024longvila} scales context via multimodal sequence parallelism. However, information discarded during compression cannot be recovered for subsequent generation. In contrast, \ours{} maintains a compact visual buffer for direct grounding while accumulating extended temporal context into an external recurrent state.

\noindent\textbf{Video memory banks.}
To maintain historical context beyond the immediate window, token-based memory banks accumulate and cluster visual features across extended horizons. MA-LMM \citep{he2024malmm} accumulates online features in memory banks, while MovieChat \citep{song2024moviechat} separates tokens into short- and long-term stores inspired by the Atkinson--Shiffrin model. VideoLLaMB \citep{wang2025videollamb} propagates recurrent memory tokens across segmented scenes, while ReWind \citep{diko2025rewind} leverages instruction-conditioned memory to retrieve salient frames for reasoning. To preserve streaming details, Flash-VStream \citep{zhang2025flashvstream} and VideoChat-Online \citep{qian2025videochatonline} incorporate multi-resolution augmentation buffers. However, these methods expose memory by appending auxiliary tokens directly to the prompt or relying on backbone-coupled bridges, which exhaust sequence capacity. \ours{} avoids sequence overhead by keeping its recurrent state outside the prompt and accessing it via additive prefix steering.

\noindent\textbf{KV-cache management.}
Managing intermediate attention key-value (KV) states offers a token-free alternative. Video-XL-2 \citep{qin2025videoxl2} and InfiniPot-V \citep{kim2025infinipotv} compress caches via task-aware sparsification or value-norm pruning. For long-term retrieval, ReKV \citep{di2025rekv} offloads chunked KV states to host memory, while StreamMem \citep{he2025streammem} aggregates incoming caches into query-agnostic fixed buffers. While effective, these methods operate directly on high-dimensional model-internal activations, tying memory management to backbone-specific attention representations and cache formats. \ours{} constructs an external recurrent state directly from visual representations and modulates prefix K/V states exclusively during prefill.

\noindent\textbf{Fast weights and associative memory.}
Fast-weight models and linear Transformers maintain context within dynamically updated parameters \citep{schmidhuber1992fastweights,katharopoulos2020transformers,schlag2021linear,ramsauer2021hopfield}. Recent systems adapt this for long sequences: Titans \citep{behrouz2025titans} and StreamTTT \citep{chen2026streamttt} employ test-time training to optimize parameters online, introducing non-trivial latency. Closest in interface, $\delta$-mem \citep{lei2026deltamem} generates low-rank attention corrections during autoregressive generation. \ours{} departs from these designs by employing closed-form, multi-factor gated outer-product writes to bypass online optimization, and applying modulations exclusively during prefill to eliminate recurrent computation during generation.
\section{Prefix-Steered Recurrent Memory}
\label{sec:method}

\begin{figure}[t]
  \vspace{-1em}
  \centering
  \includegraphics[width=\linewidth]{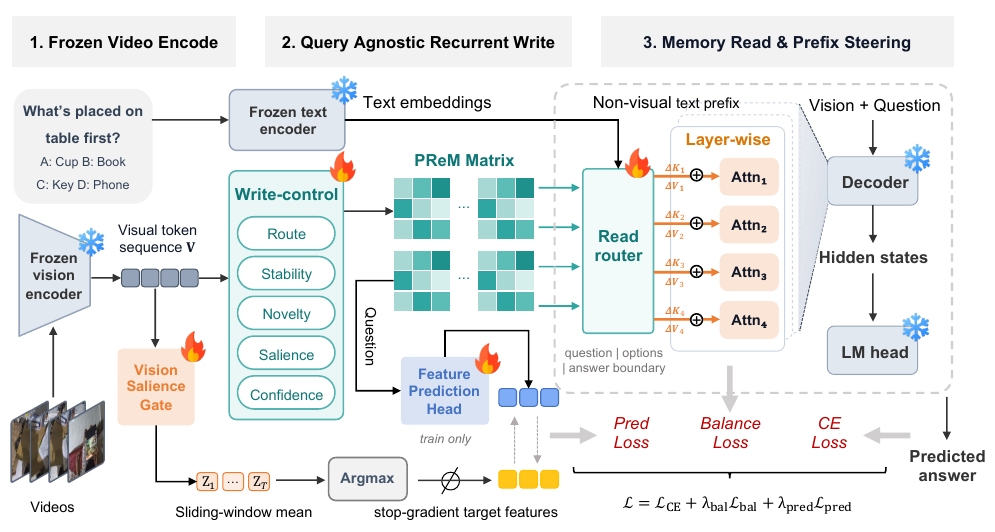}
  \caption{\ours{} architecture and supervision. A query-agnostic gated residual rule updates a $K$-slot recurrent state $S$. During QA, prefix representations route memory reads into additive K/V steering. The visual buffer remains standard input; state $S$ operates outside the prompt with zero tokens.}
  \label{fig:framework}
  \vspace{-1em}
\end{figure}

\noindent\textbf{Problem formulation and overview.}
Given an extended video $\mathcal{V} = \{v_t\}_{t=1}^T$ of $T$ frames and a textual query $\mathcal{Q}$, long-video understanding aims to generate an answer $A$ conditioned on both modalities. \ours{} addresses this by decoupling temporal retention from sequence capacity via a dual-pathway interface (Figure~\ref{fig:framework}). The primary pathway feeds a bounded visual buffer $\mathcal{V}_B \subset \mathcal{V}$ containing $B$ frames into a frozen VLM backbone for direct visual grounding. Concurrently, a recurrent writer ingests the full frame sequence directly from visual encoder representations, updating an external, constant-footprint multi-slot associative state $S$ while bypassing the LLM backbone during ingestion. At query time, a question-conditioned readout projects $S$ into additive steering vectors $d_i^{k,(\ell)}, d_i^{v,(\ell)}$ to modulate non-visual prefix K/V representations during prefill. This mechanism steers backbone attention over the visual buffer without adding sequence tokens or fine-tuning backbone weights.

\subsection{Query-Agnostic Recurrent Write}
\label{sec:write}

Let $E(v_t)_p \in \mathbb{R}^{d_{\mathrm{enc}}}$ denote the feature from the frozen vision encoder for frame $v_t \in \mathcal{V}$ at spatial patch $p \in \{1, \dots, P_t\}$, where $P_t$ is the number of patch tokens at step $t$. We obtain the frame-level visual feature $x_t$ through spatial mean-pooling and project it into address and content spaces:
\begin{equation}
  x_t=\frac{1}{P_t}\sum_{p=1}^{P_t}E(v_t)_p,\qquad
  k_t=W_k\operatorname{LN}(x_t),\quad
  u_t=W_v\operatorname{LN}(x_t),
  \label{eq:visual-feature}
\end{equation}
where $\operatorname{LN}(\cdot)$ denotes layer normalization. The learnable matrices $W_k \in \mathbb{R}^{d_k \times d_{\mathrm{enc}}}$ and $W_v \in \mathbb{R}^{d_v \times d_{\mathrm{enc}}}$ project $x_t$ into address and content spaces, respectively.
Writer features follow the visual-to-text projection ($d_{\mathrm{enc}}=d_{\mathrm{model}}$); spatial patch tokens remain available in the visual buffer.

Given the sequence $\{x_t\}_{t=1}^T$, the recurrent state maintains $K$ matrix slots $S_m\in\mathbb{R}^{d_v\times d_k}$ and slot confidences $c_m \in [0, 1]$, where $d_k$ and $d_v$ denote key and value dimensions. A learnable address transform $A_m \in \mathbb{R}^{d_k \times d_k}$, initialized near identity, parameterizes each slot:
\begin{equation}
  \kappa_{t,m}=\operatorname{norm}(A_m k_t),\qquad
  e_{t,m}=u_t-S_m\kappa_{t,m},
  \label{eq:write-residual}
\end{equation}
where $\operatorname{norm}(v) = v / \max(\lVert v\rVert_2, \epsilon_{\mathrm{norm}})$ denotes $\ell_2$ normalization with numerical stabilizer $\epsilon_{\mathrm{norm}}=10^{-12}$. The vector $\kappa_{t,m} \in \mathbb{R}^{d_k}$ is the unit-norm write address, and residual $e_{t,m} \in \mathbb{R}^{d_v}$ measures the discrepancy between projected value $u_t$ and slot prediction $S_m\kappa_{t,m}$.

\begin{figure}[t]
  \centering
  \includegraphics[width=\linewidth]{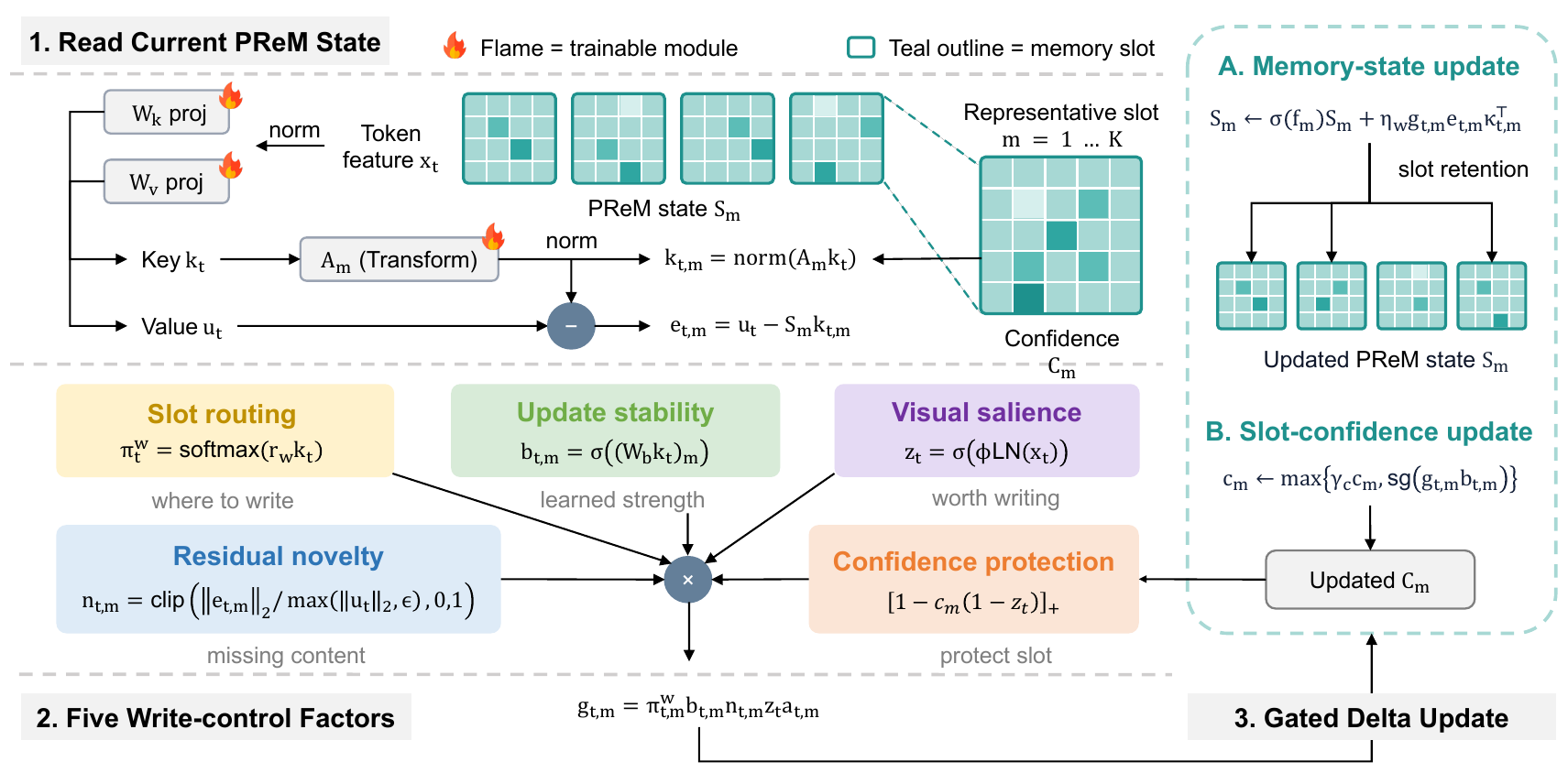}
  \caption{Recurrent write update. Subtracting slot predictions from visual projections yields associative residuals, modulated by multi-factor controls to update decayed memory and slot confidences.}
  \label{fig:write}
  \vspace{-1em}
\end{figure}

As illustrated in Figure~\ref{fig:write}, five bounded gates regulate memory updates over extended horizons:
\begin{align}
  \pi^w_t &= \operatorname{softmax}(r_w k_t),
  & b_{t,m} &= \sigma((W_bk_t)_m), \nonumber\\
  n_{t,m} &= \operatorname{clip}\!\left(
    \frac{\lVert e_{t,m}\rVert_2}{\max(\lVert u_t\rVert_2,\epsilon)},0,1\right),
  & z_t &= \sigma(\phi(\operatorname{LN}(x_t))), \nonumber\\[-2pt]
  a_{t,m} &= [1-c_m(1-z_t)]_+,&
  g_{t,m} &= \pi^w_{t,m} b_{t,m} n_{t,m} z_t a_{t,m},
  \label{eq:gate}
\end{align}
where $\sigma(\cdot)$ is sigmoid, $[\cdot]_+=\max(0,\cdot)$ is ReLU, and $\epsilon=10^{-6}$ is a numerical stabilizer. Affine gate biases are implicit. The five gates regulate updates as follows:
\textbf{(1)~Slot routing} ($\pi_t^w\in\Delta^K$) maps $k_t$ via $r_w\in\mathbb{R}^{K\times d_k}$ for semantic slot specialization;
\textbf{(2)~Update stability} ($b_{t,m}\in(0,1)$) uses $W_b\in\mathbb{R}^{K\times d_k}$ to scale steps and suppress perturbations;
\textbf{(3)~Residual novelty} ($n_{t,m}\in[0,1]$) suppresses redundant writes when $e_{t,m}\approx0$;
\textbf{(4)~Visual salience} ($z_t\in(0,1)$) downweights background frames via $\phi$, a $d_{\mathrm{model}}\!\to\!\lfloor d_{\mathrm{model}}/4\rfloor\!\to\!1$ GELU MLP; and
\textbf{(5)~Confidence protection} ($a_{t,m}\in[0,1]$) protects confident slots ($c_m\to1$) from low-salience writes ($z_t\to0$).

The gate $g_{t,m}$ modulates the outer-product update for each slot matrix $S_m$ and its confidence $c_m$:
\begin{equation}
  S_m\leftarrow \sigma(f_m)S_m+\eta_w g_{t,m}e_{t,m}\kappa_{t,m}^{\top},
  \qquad
  c_m\leftarrow\max\{\gamma_c c_m,\operatorname{sg}(g_{t,m}b_{t,m})\},
  \label{eq:update}
\end{equation}
where $f_m \in \mathbb{R}$ is a learnable slot-wise forget logit, and $\sigma(f_m)$ controls gradual memory decay. The fixed factor $\eta_w > 0$ sets the write step size, while $\gamma_c \in (0, 1)$ sets the exponential confidence decay. The stop-gradient operator $\operatorname{sg}(\cdot)$ blocks confidence gradients. The recurrent state occupies $\mathcal{O}(K d_v d_k)$ memory (${\approx}256$\,KiB for $K{=}4$ and $d_v{=}d_k{=}128$), remaining constant regardless of video duration.

\vspace{-1em}

\subsection{Question-Conditioned Read and Prefix Steering}
\label{sec:read}

Because the recurrent write update is query-agnostic (Eqs.~\ref{eq:visual-feature}--\ref{eq:update}), the multi-slot state $S = \{S_m\}_{m=1}^K$ supports repeated queries after a single video ingestion pass. At query time, we average embeddings of existing non-visual prefix tokens (instructions, query $\mathcal{Q}$, options, and prompt boundaries) into a global question representation $q \in \mathbb{R}^{d_{\mathrm{model}}}$. To accommodate different abstraction levels across depth, we partition the $L$ backbone layers into $G$ contiguous groups, where layer $\ell \in \{1, \dots, L\}$ maps to group $g(\ell) = \lfloor (\ell - 1) G / L \rfloor \in \{0, \dots, G-1\}$.

At layer $\ell$ and prefix position $i$, the hidden state $h_i^{(\ell)} \in \mathbb{R}^{d_{\mathrm{model}}}$ is conditioned on $q$ and projected to the normalized read address $\xi_{i,m}^{(\ell)} \in \mathbb{R}^{d_k}$, retrieving value $r_{i,m}^{(\ell)} \in \mathbb{R}^{d_v}$ from slot $m$:
\begin{equation}
  \begin{aligned}
    \tilde h_i^{(\ell)}&=\operatorname{LN}(h_i^{(\ell)})+W_c\operatorname{LN}(q),\quad & q_i^{(\ell)}&=W_r\operatorname{LN}(\tilde h_i^{(\ell)}),\\
    \xi_{i,m}^{(\ell)}&=\operatorname{norm}(A_mq_i^{(\ell)}),\quad & r_{i,m}^{(\ell)}&=S_m\xi_{i,m}^{(\ell)},
  \end{aligned}
  \label{eq:read}
\end{equation}
where $W_c \in \mathbb{R}^{d_{\mathrm{model}} \times d_{\mathrm{model}}}$ is a conditioning projection, $W_r \in \mathbb{R}^{d_k \times d_{\mathrm{model}}}$ maps conditioned features to query space, and $A_m$ is the slot address transform from Eq.~\ref{eq:write-residual}.

A router $R: \mathbb{R}^{3d_v} \to \mathbb{R}$ evaluates conditioned query $q_i^{(\ell)}$, retrieved value $r_{i,m}^{(\ell)}$, and their Hadamard product, producing relevance logits $s_{i,m}^{(\ell)}$ that normalize into routing weights $\rho^{r,(\ell)}_{i,:} \in \Delta^K$:
\begin{equation}
  s_{i,m}^{(\ell)}=R([W_{qv}q_i^{(\ell)}; r_{i,m}^{(\ell)}; W_{qv}q_i^{(\ell)}\odot r_{i,m}^{(\ell)}]),\qquad
  \rho^{r,(\ell)}_{i,:}=\operatorname{softmax}_m(s_{i,m}^{(\ell)}/\tau),
  \label{eq:read-router}
\end{equation}
where $[\cdot;\cdot]$ denotes concatenation, $\odot$ the Hadamard product, and $\tau>0$ the routing temperature. $W_{qv}\in\mathbb{R}^{d_v\times d_k}$ aligns dimensions; $R$ is a two-layer GELU router (Table~\ref{tab:repro-constants}). Group-specific heads $H_{g(\ell),m}^a\in\mathbb{R}^{d_{\mathrm{model}}\times d_v}$ ($a\in\{k,v\}$) map weighted reads to steering vectors $d_i^{a,(\ell)}\in\mathbb{R}^{d_{\mathrm{model}}}$:
\begin{equation}
  d_i^{a,(\ell)}=\frac{\alpha_{\mathrm{eff}}}{K}
  \sum_{m=1}^{K}\rho^{r,(\ell)}_{i,m}H_{g(\ell),m}^{a}r_{i,m}^{(\ell)},\qquad
  \alpha_{\mathrm{eff}}=\exp(\hat\alpha)\alpha_{\mathrm{run}},
  \label{eq:correction}
\end{equation}
where $\hat\alpha \in \mathbb{R}$ is a learnable scale logit and $\alpha_{\mathrm{run}}$ is an inference-time scale. The steering vectors define a hidden-state perturbation before the frozen K/V projections:
\begin{equation}
  K_i^{(\ell)}=W_K^{(\ell)}(h_i^{(\ell)}+d_i^{k,(\ell)}),
  \qquad
  V_i^{(\ell)}=W_V^{(\ell)}(h_i^{(\ell)}+d_i^{v,(\ell)}),
  \label{eq:modulation}
\end{equation}
where $W_K^{(\ell)},W_V^{(\ell)}$ are frozen projections (biases omitted). The perturbations induce additive shifts $\Delta K=W_Kd^k$ and $\Delta V=W_Vd^v$ at the projection outputs, followed by the backbone's native attention transforms (Appendix~\ref{sec:appendix-interface}).

\paragraph{Prefill-only steering.}
Modulation applies exclusively to non-visual prompt prefixes (e.g., instructions, questions $\mathcal{Q}$) during prefill. Modulating prefix keys ($\Delta K$) reshapes prompt-level attention addressing, while values ($\Delta V$) inject distilled long-horizon evidence that causally guides subsequent layers. Once prefill completes, these representations persist in the standard KV cache, so autoregressive generation inherits full-horizon guidance without recurrent operations. Appendices~\ref{sec:appendix-interface} and~\ref{sec:appendix-state} detail the attention interface and initialization.

\subsection{Training and Inference Protocols}
\label{sec:training_inference}

During training, videos are sampled uniformly at a fixed frame rate. The writer ingests an incoming sequence of $T$ frames to update the recurrent state $S$, while the decoder prompt draws a visual buffer $\mathcal{V}_B$ of $B$ frames from the same stream. We optimize solely \ours{} parameters by minimizing:
\begin{equation}
  \mathcal L=\mathcal L_{\mathrm{CE}}
  +\lambda_{\mathrm{bal}}\mathcal L_{\mathrm{bal}}
  +\lambda_{\mathrm{pred}}\mathcal L_{\mathrm{pred}},
  \label{eq:loss}
\end{equation}
where $\mathcal L_{\mathrm{CE}}$ is the cross-entropy loss, and $\lambda_{\mathrm{bal}}, \lambda_{\mathrm{pred}}$ weight auxiliary objectives. The load-balancing term $\mathcal L_{\mathrm{bal}}=\sum_m(\bar{\rho}_m-1/K)^2$ averages read-routing weights $\bar{\rho}_m$ over valid prefixes. The prediction loss masks a high-salience temporal span during a second pass and reconstructs its normalized value features from the held-out state, conditioned on the question and temporal position. Appendix~\ref{sec:appendix-prediction} details the head, shapes, target, and loss. During inference, \ours{} applies the same update and modulation interfaces (Eqs.~\ref{eq:update} and~\ref{eq:modulation}) in offline and streaming end-of-stream modes: offline ingestion processes the full stream once, whereas streaming carries state $(S,c)$ across chunks and answers after the video ends, both predicting $A$ via one-token constrained decoding.

\section{Experiments}
\label{sec:experiments}

We evaluate \ours{} across four questions:
\begin{itemize}[leftmargin=*, itemsep=0pt, topsep=0pt]
  \item \textbf{RQ1 (Performance):} How does \ours{} compare with long-video and streaming methods? (\S\ref{sec:main-results})
  \item \textbf{RQ2 (Scaling):} Do the gains hold across visual budgets, durations, and backbones? (\S\ref{sec:budget-results})
  \item \textbf{RQ3 (Ablations):} What are the contributions of prefix steering and write gating, and how stable is the design under training horizon and hyperparameters? (\S\ref{sec:ablation})
  \item \textbf{RQ4 (Efficiency):} Where do efficiency and task gains originate under matched settings? (\S\ref{sec:analysis})
\end{itemize}

\subsection{Experimental Setup}
\label{sec:setup}

\paragraph{Models and baselines.}
We instantiate \ours{} on three frozen backbones: Qwen2.5-VL-3B, Qwen3-VL-8B, and LLaVA-Video-7B \citep{bai2025qwen25vl,bai2025qwen3vl,zhang2024llavavideo}. Per Table~\ref{tab:main}, baselines include: (1)~\textbf{specialized architectures} spanning compression (LLaMA-VID, VideoLLaMA2, LLaVA-Video, LongVILA, LongVU) and streaming memory (MovieChat, StreamChat, ReKV, Flash-VStream) under native budgets; and (2)~\textbf{generalist base VLMs} (Qwen2.5-VL-3B, Qwen3-VL-8B). We train only \ours{} for one epoch on a LLaVA-Video-178K subset (6{,}430 videos, 17{,}849 QA pairs, $<1.4\%$ of its QA pairs; Table~\ref{tab:train-data}). Trainable parameters total 9.07M (Qwen2.5) and 27.48M (Qwen3) with $d_v{=}d_k{=}128$ (${\approx}0.24\%$--$0.30\%$). The standard configuration uses $K{=}4$; Table~\ref{tab:same-backbone} uses $K{=}1$ for the controlled comparison. Training samples 1\,FPS with write cap $T{=}64$ and buffer $B$.

\vspace{-0.5em}
\paragraph{Evaluation protocol.}
Evaluation spans six benchmarks: LongVideoBench (LVB) \citep{wu2024longvideobench}, MLVU \citep{zhou2024mlvu}, Video-MME (VMME) \citep{fu2025videomme}, EgoSchema (EgoS) \citep{mangalam2023egoschema}, MVBench (MVB) \citep{li2023mvbench}, and LVBench (LVBN) \citep{wang2024lvbench}. Models predict via one-token constrained decoding; we report per-benchmark accuracy and macro \textbf{Avg}. Per \S\ref{sec:training_inference}, evaluation ingests a question-agnostic 1\,FPS stream ($T{\le}240$). Matched controls fix the backbone, prompt template, decoder budget $B \in \{16, 32, 64, 90\}$, resolution cap, and decoding rule; Base ($B{=}240$) is a high-budget reference. Offline mode writes the stream once; streaming mode ingests two-frame chunks carrying state $(S,c)$ and answers at stream end (detailed in Appendix~\ref{sec:appendix}).

\vspace{-0.5em}
\subsection{RQ1: Performance Comparison}
\label{sec:main-results}

\begin{table}[!t]
\vspace{-1em}
\caption{\tabtitle{Public-benchmark comparison} (\%). External systems follow official configurations (Appendix~\ref{sec:appendix}). Base VLMs and \ours{} use matched $B{=}90$ on frozen backbones. Subscripts denote paired gains over Base. Bold marks top performance within each regime.}
\label{tab:main}
\centering
\scriptsize
\setlength{\tabcolsep}{2.5pt}
\renewcommand{\arraystretch}{0.95}
\resizebox{\linewidth}{!}{%
\begin{tabular}{@{}llccccccc@{}}
\toprule
\tabhead{Method (Backbone)} & \tabhead{Tuning / Budget} & \tabhead{LVB} & \tabhead{MLVU} & \tabhead{VMME} & \tabhead{EgoS} & \tabhead{MVB} & \tabhead{LVBN} & \tabhead{Avg} \\
\midrule
\multicolumn{9}{@{}l}{\textbf{Offline Ingestion} \quad \textit{(Specialized long-video architectures)}} \\
LLaMA-VID (Vicuna-7B) & Full-FT (1 FPS) & 28.27 & 35.26 & 30.11 & 27.20 & 32.78 & 25.89 & 29.92 \\
VideoLLaMA2 (Qwen2-7B) & Full-FT ($B{=}16$) & 48.54 & 56.92 & 54.59 & 61.20 & 55.43 & 36.73 & 52.24 \\
LLaVA-Video (Qwen2-7B) & SFT ($B{=}64$) & 59.39 & 70.02 & 63.33 & 58.20 & 59.60 & 42.41 & 58.82 \\
LongVILA (Qwen2-7B) & Long-FT ($B{=}32/256$) & 55.27 & 65.33 & 60.11 & 72.20 & 64.65 & 40.99 & 59.76 \\
LongVU (Qwen2-7B) & SFT (0.5 FPS) & 52.80 & 65.06 & 56.30 & 70.00 & 64.43 & 37.70 & 57.72 \\
StreamChat (LongVA-7B) & Train-free (20\%) & 46.00 & 54.30 & 46.15 & 47.20 & 45.73 & 35.31 & 45.78 \\
ReKV (LLaVA-OV-7B) & Train-free (0.5 FPS) & 56.17 & 67.03 & 57.67 & 60.80 & 55.43 & 39.77 & 56.15 \\
Flash-VStream (Qwen2-VL-7B) & LoRA ($B{=}240$) & 52.06 & 62.71 & 58.19 & 69.60 & 60.42 & 38.22 & 56.87 \\
\midrule
\multicolumn{9}{@{}l}{\textit{Generalist base VLMs \& \ours{} (matched $B{=}90$)}} \\
Qwen2.5-VL-3B & Frozen ($B{=}90$) & 52.13 & 63.45 & 61.26 & 61.20 & 57.92 & 40.41 & 56.06 \\
\rowcolor{asmlight}\ours{} (Qwen2.5-VL-3B) & Plug-in ($B{=}90$) & 55.27$_{\scriptscriptstyle +3.14}$ & 65.47$_{\scriptscriptstyle +2.02}$ & 61.30$_{\scriptscriptstyle +0.04}$ & 61.20 & 61.35$_{\scriptscriptstyle +3.43}$ & 43.12$_{\scriptscriptstyle +2.71}$ & 57.95$_{\scriptscriptstyle +1.89}$ \\
Qwen3-VL-8B & Frozen ($B{=}90$) & 60.88 & 70.94 & 67.56 & 72.60 & 65.90 & 45.32 & 63.87 \\
\rowcolor{asmlight}\ours{} (Qwen3-VL-8B) & Plug-in ($B{=}90$) & \textbf{63.05}$_{\scriptscriptstyle +2.17}$ & \textbf{71.22}$_{\scriptscriptstyle +0.28}$ & \textbf{67.87}$_{\scriptscriptstyle +0.31}$ & \textbf{74.20}$_{\scriptscriptstyle +1.60}$ & \textbf{67.10}$_{\scriptscriptstyle +1.20}$ & \textbf{45.76}$_{\scriptscriptstyle +0.44}$ & \textbf{64.87}$_{\scriptscriptstyle +1.00}$ \\
\midrule[\heavyrulewidth]
\multicolumn{9}{@{}l}{\textbf{Streaming End-of-Stream} \quad \textit{(Chronological Ingestion, native budgets)}} \\
MovieChat (Vicuna-7B) & Train-free ($128{\times}8$) & 21.17 & 28.55 & 26.44 & 20.40 & 29.38 & 23.56 & 24.92 \\
StreamChat (LongVA-7B) & Train-free (20\%) & 46.00 & 53.20 & 46.63 & 49.00 & 45.68 & 34.41 & 45.82 \\
ReKV (LLaVA-OV-7B) & Train-free (0.5 FPS) & 56.17 & 67.03 & 57.67 & 60.80 & 55.43 & 39.83 & 56.16 \\
Flash-VStream (Qwen2-VL-7B) & LoRA ($B{=}240$) & 52.28 & 60.55 & 55.89 & 70.40 & 57.62 & 37.19 & 55.66 \\
\midrule
\multicolumn{9}{@{}l}{\textit{Generalist base VLMs \& \ours{} (matched $B{=}90$)}} \\
Qwen2.5-VL-3B & Frozen ($B{=}90$) & 54.75 & 63.59 & 59.63 & 61.40 & 61.20 & 41.77 & 57.06 \\
\rowcolor{asmlight}\ours{} (Qwen2.5-VL-3B) & Plug-in ($B{=}90$) & 56.54$_{\scriptscriptstyle +1.79}$ & 65.52$_{\scriptscriptstyle +1.93}$ & 61.30$_{\scriptscriptstyle +1.67}$ & 61.20$_{\scriptscriptstyle -0.20}$ & 61.40$_{\scriptscriptstyle +0.20}$ & 42.93$_{\scriptscriptstyle +1.16}$ & 58.15$_{\scriptscriptstyle +1.09}$ \\
Qwen3-VL-8B & Frozen ($B{=}90$) & 61.71 & 69.75 & 67.22 & 72.40 & 66.05 & 42.61 & 63.29 \\
\rowcolor{asmlight}\ours{} (Qwen3-VL-8B) & Plug-in ($B{=}90$) & \textbf{62.15}$_{\scriptscriptstyle +0.44}$ & \textbf{71.22}$_{\scriptscriptstyle +1.47}$ & \textbf{67.87}$_{\scriptscriptstyle +0.65}$ & \textbf{74.20}$_{\scriptscriptstyle +1.80}$ & \textbf{67.13}$_{\scriptscriptstyle +1.08}$ & \textbf{45.76}$_{\scriptscriptstyle +3.15}$ & \textbf{64.72}$_{\scriptscriptstyle +1.43}$ \\
\bottomrule
\end{tabular}}
\end{table}
\vspace{-0.5em}

Table~\ref{tab:main} provides system-level context from specialized long-video systems at native budgets and measures gains against matched-budget frozen base VLMs ($B{=}90$). \ours{} remains competitive with specialized long-video systems under their native configurations. Across both backbones, \ours{} provides consistent gains over paired frozen bases: offline Avg rises by $+1.89\%$ (3B) and $+1.00\%$ (8B); streaming Avg rises by $+1.09\%$ (3B) and $+1.43\%$ (8B). In streaming, \ours{} (Qwen3-VL-8B) reaches 64.72\% Avg while training only 27.5M parameters ($<0.30\%$).

\vspace{-0.5em}
\subsection{RQ2: Budget Scaling and Generalization}
\label{sec:budget-results}

\begin{figure}[b!]
  \centering
  \includegraphics[width=\textwidth]{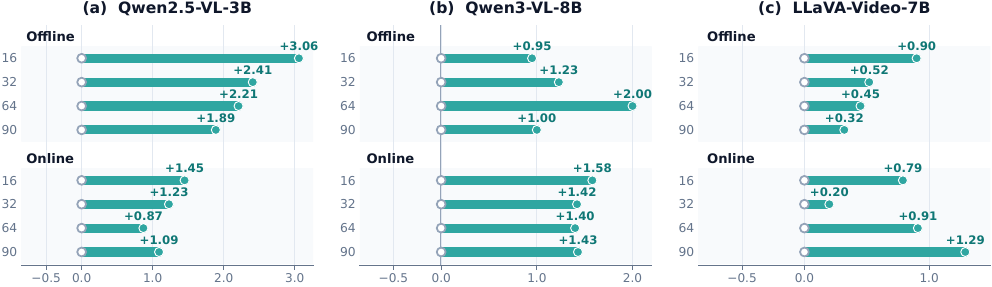}
  \caption{Horizontal axis: $\ours{}-\mathrm{Base}$ Avg gain; vertical axis: visual budget $B \in \{16, 32, 64, 90\}$. Open/filled markers denote Base/\ours{}. Full breakdowns appear in Appendix Tables~\ref{tab:budget} and~\ref{tab:budget-llava}.}
  \label{fig:budget-gains}
  \vspace{-1em}
\end{figure}

\paragraph{Budget scaling and backbone transfer.}
Figure~\ref{fig:budget-gains} shows \ours{} outperforming frozen baselines across budgets $B \in \{16, 32, 64, 90\}$. On Qwen2.5-VL-3B, gains peak under temporal sparsity ($B{=}16$): offline Avg rises by \textbf{+3.06\%} (50.39\% to 53.45\%) and streaming Avg by \textbf{+1.45\%}. On Qwen3-VL-8B, steady gains reach \textbf{+2.00\%} offline ($B{=}64$) and \textbf{+1.58\%} in streaming ($B{=}16$). At $B{=}90$, margins persist ($+1.89\%$ and $+1.00\%$). Gains transfer to LLaVA-Video-7B ($+0.90\%$ offline at $B{=}16$; $+0.20$--$+1.29\%$ streaming; Table~\ref{tab:budget-llava}), confirming generality.

\begin{figure}[t]
  \centering
  \includegraphics[width=\textwidth]{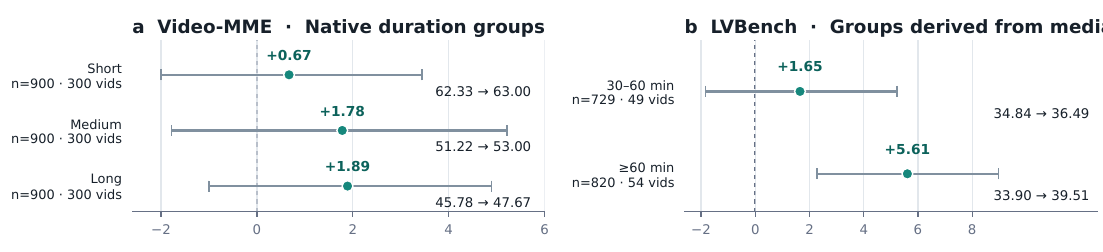}
  \caption{Duration-stratified $\ours{}-\mathrm{Base}$ gains on Qwen2.5-VL-3B ($B{=}16$, $K{=}4$, $T_{\mathrm{mem}}{=}16$; Appendix~\ref{sec:appendix-diagnostics}). Error bars: 95\% paired bootstrap CIs (20k); labels: $\mathrm{Base}\to\ours{}$ accuracy.}
  \label{fig:duration-gains}
\end{figure}

\paragraph{Generalization across native video durations.}
Figure~\ref{fig:duration-gains} stratifies the four-slot diagnostic run by native video duration. On Video-MME, gains increase monotonically with video length: $+0.67\%$ on Short, $+1.78\%$ on Medium, and $+1.89\%$ on Long videos. On LVBench, the advantage widens to \textbf{+5.61\%} on videos of at least 60 minutes, with the 95\% bootstrap CI strictly above zero. This diagnostic run shows its largest gain in the longest LVBench group.

\subsection{RQ3: Ablations and Training Robustness}
\label{sec:ablation}

\begin{figure}[h]
  \centering
  \includegraphics[width=\textwidth]{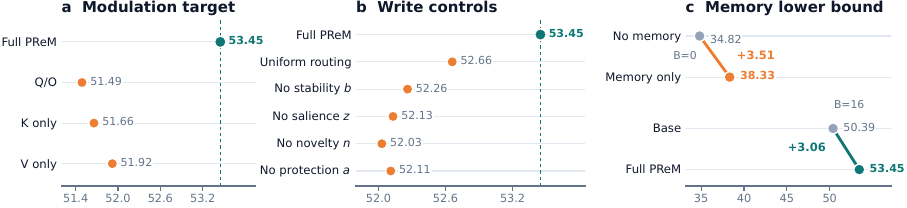}
  \caption{Diagnostic ablations on Qwen2.5-VL-3B (offline $B{=}16$). Horizontal axis: Avg accuracy (\%). \textbf{(a)} Steering target interfaces. \textbf{(b)} Write gating terms against Full \ours{} (dashed line). \textbf{(c)} Zero-buffer probe ($B{=}0$) isolating recurrent retention against $B{=}16$ Base reference.}
  \label{fig:ablation}
  \vspace{-1em}
\end{figure}

\paragraph{Diagnostic ablations.}
Figure~\ref{fig:ablation} reports component ablations on Qwen2.5-VL-3B ($B{=}16$) under a shared experimental setup. Full \ours{} reaches 53.45\% Avg; K-only and V-only variants reach 51.66\% and 51.92\%, while write-gate variants span 52.03\%--52.66\%. The zero-buffer probe ($B{=}0$) reaches 38.33\%, $+3.51$ above the vision-free Base.

\paragraph{Training efficiency and horizon generalization.}
Figure~\ref{fig:efficiency-horizon}a--b sweeps writer cap $T$ against full-horizon evaluation ($T{=}240, B{=}16$). Even at $T{=}16$, enabling $S$ boosts offline Avg by $+2.17$ over the clean Base, reaching 52.56\%. The selected $T{=}64$ checkpoint reaches 53.45\% offline and 53.46\% streaming at $B{=}16$ (Table~\ref{tab:budget}), while using 73.3\% fewer writer steps than $T{=}240$. Thus \ours{} learns effective recurrent dynamics from short horizons.

\paragraph{Sensitivity.}
We evaluate 36 configurations and a clean Base reference across eight hyperparameters (Table~\ref{tab:sen}; Figure~\ref{fig:sensitivity-app}). Training and capacity sweeps yield 51.29\%--52.76\% Avg, with the slot sweep peaking at $K{=}4$. Inference scales $\alpha_{\mathrm{run}}{=}0.5/1.0$ yield 52.26\%/51.74\%; larger scales $1.5/2.0$ reduce Avg to 48.79\%/43.56\%. Multi-seed results appear in Appendix~\ref{sec:appendix-seeds}.

\subsection{RQ4: Efficiency and Task Analysis}
\label{sec:analysis}

\paragraph{Controlled streaming and task advantages.}
Table~\ref{tab:same-backbone} compares the three memory paradigms from Sec.~1 on Qwen2.5-VL-3B at $B{=}16$ (details in Appendix~\ref{sec:appendix-same-backbone}). Parameter-matched \ours{} ($K{=}1$) achieves top Avg (53.46\%), outperforming Base (52.01\%), LoRA (51.21\%), Token-Readout ($K{=}1$, 51.33\%), and InfiniPot-V (51.73\%). InfiniPot-V reaches 61.08\% on MVB and 54.48\% on VMME, while its EgoSchema accuracy is 50.20\% ($4.20$ points below Base), with an 84.24\,MiB footprint and 17.44\,s ingest latency. In contrast, \ours{} retains long-horizon evidence (EgoS 54.60\%, LVBN 38.48\%) with a 0.063\,MiB state and zero extra prompt tokens, cutting ingestion time to 7.81\,s.

\begin{figure*}[h]
  \centering
  \includegraphics[width=\textwidth]{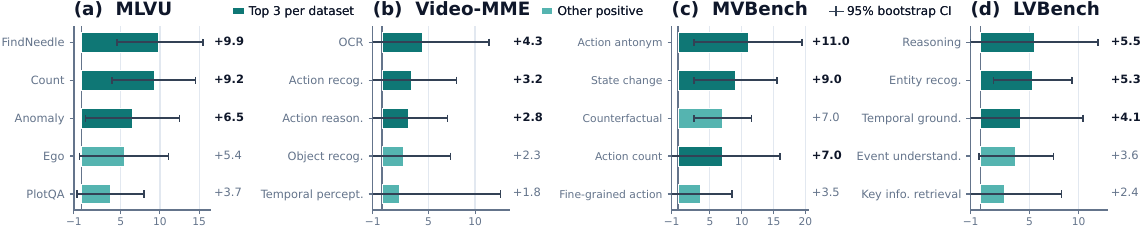}
  \caption{Highest-gain task categories on Qwen2.5-VL-3B (offline $B{=}16$). Horizontal axis: paired $\ours{}-\text{Base}$ accuracy difference (\%) with 95\% bootstrap CIs. The complete category breakdown is provided in Appendix Figure~\ref{fig:advantage-cases-full}.}
  \label{fig:advantage-cases}
\end{figure*}

\vspace{-1em}
\begin{table*}[h]
\vspace{-1em}
\caption{\tabtitle{Controlled streaming mechanism comparison on Qwen2.5-VL-3B.} Evaluated under matched input streams ($T{=}240, B{=}16$) and parameter budgets. Bold and underline mark top and second-best results per benchmark.}
\label{tab:same-backbone}
\centering
\scriptsize
\setlength{\tabcolsep}{1.35pt}
\renewcommand{\arraystretch}{1.1}
\begin{tabular*}{\textwidth}{@{\extracolsep{\fill}}lcccccccccccc@{}}
\toprule
& & \multicolumn{7}{c}{\textbf{Benchmark Accuracy (\%)}} & \multicolumn{4}{c}{\textbf{Resource Accounting}} \\
\cmidrule(lr){3-9} \cmidrule(lr){10-13}
Method & \makecell{Params\\(M)} & LVB & MLVU & VMME & EgoS & MVB & LVBN & \textbf{Avg} & \makecell{State\\(MiB / Tok.)} & \makecell{VRAM\\(GiB)} & \makecell{Ingest\\(s)} & \makecell{Answer\\(s)} \\
\midrule
Base (Frozen, $B{=}16$)                       & 0    & \underline{52.73} & 55.68 & 53.89 & \underline{54.40} & 59.98 & 35.38 & \underline{52.01} & 0.000 / 0   & 7.43 &  5.64 & 0.30 \\
\quad + Parameter Fine-tuning (LoRA, $r{=}20$) & 9.22 & 49.36 & 55.45 & 53.04 & \textbf{54.60} & 57.98 & \underline{36.86} & 51.21 & 0.000 / 0   & 7.46 &  5.64 & 0.30 \\
\quad + Token-Readout (Memory Bank)            & 9.07 & 51.23 & 55.59 & 53.48 & 52.80 & 58.70 & 36.15 & 51.33 & 0.063 / +10 & 7.48 &  9.20 & 0.35 \\
\quad + InfiniPot-V (KV-Cache Management)      & 0    & 52.06 & \underline{57.10} & \textbf{54.48} & 50.20 & \textbf{61.08} & 35.44 & 51.73 & 84.236 / 0  & 7.99 & 17.44 & 0.32 \\
\rowcolor{gray!10}
\quad + \ours{} (Prefix Recurrent Memory)      & 9.07 & \textbf{53.55} & \textbf{59.68} & \underline{54.11} & \textbf{54.60} & \underline{60.33} & \textbf{38.48} & \textbf{53.46} & 0.063 / 0 & 7.46 & 7.81 & 0.38 \\
\bottomrule
\end{tabular*}
\end{table*}

\begin{figure}[!t]
  \centering
  \includegraphics[width=\linewidth]{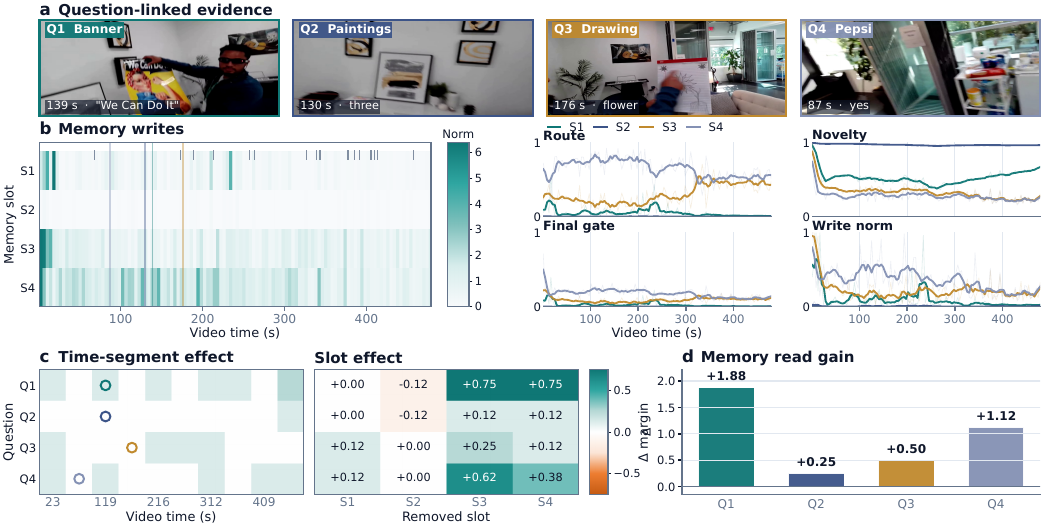}
  \caption{Qualitative write-read dynamics on EgoSchema. \textbf{(a)} Evidence frames. \textbf{(b)} Write intensity and gating. \textbf{(c)} Margin drop $m_{\mathrm{full}}-m_{\mathrm{ablated}}$ under temporal/slot removal, where $m$ is the gold-answer logit minus the largest alternative logit. \textbf{(d)} Readout gain $m_{\mathrm{on}}-m_{\mathrm{off}}$.}
  \label{fig:prem-case-study}
  \vspace{-1em}
\end{figure}

\paragraph{Single-video case study.}
Figure~\ref{fig:prem-case-study} examines a four-slot diagnostic checkpoint on an EgoSchema example. For Q1 (querying banner text at 139\,s), ablating $S_3$ or $S_4$ reduces the gold margin by $0.75$, whereas ablating $S_1$ or $S_2$ causes no degradation. Enabling prefix steering shifts the prediction from D (incorrect) to B (gold), illustrating slot-specific contributions.

\begin{figure*}[h]
  \centering
  \includegraphics[width=\textwidth]{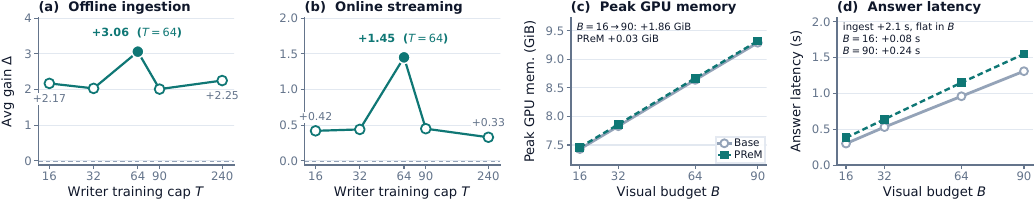}
  \caption{Training horizon and cost on Qwen2.5-VL-3B. \textbf{(a--b)} Training writer cap $T$ (evaluation $T{=}240$, $B{=}16$; Table~\ref{tab:writer-t}); dashed: zero gain; filled: selected $T{=}64$. \textbf{(c)} Peak memory tracks Base: $B$ ($16{\to}90$) costs $+1.86$\,GiB, \ours{} a constant $+0.03$\,GiB. \textbf{(d)} Answering latency closely tracks Base across visual budgets (${\le}0.24$\,s difference), while ingestion adds ${\approx}2$\,s per QA (Table~\ref{tab:efficiency}).}
  \label{fig:efficiency-horizon}
  \vspace{-1em}
\end{figure*}

\paragraph{Runtime efficiency.}
Figure~\ref{fig:efficiency-horizon}c--d summarizes efficiency: peak GPU memory increases by only 0.03\,GiB across budgets. Ingestion adds ${\approx}2$\,s, while answering latency matches Base within 0.08\,s ($B{=}16$) and 0.24\,s ($B{=}90$). Across benchmarks, 41.4\% of questions share cached $(S,c)$ states, amortizing ingestion across repeated queries.

\section{Conclusion}
\label{sec:conclusion}

\ours{} shows that a compact recurrent state, steered into a frozen backbone via prefix keys and values, achieves long-horizon video understanding without prompt tokens while tuning only a lightweight module. It accumulates evidence under visual sparsity, enabling write-once, query-many reuse without decoding recurrence. Future work includes extending \ours{} with spatially grounded memory for embodied agents \citep{anwar2024remembr}.

\clearpage

\section*{AI Use Statement}
Generative AI tools assisted with manuscript critique, methodology and experiment feedback, language editing, \LaTeX{} grammar checking, and text formatting. The authors independently made all methodological, implementation, experimental, and analytical decisions. All AI-assisted text and empirical claims were reviewed against source code, logs, and checkpoints, and the authors take full responsibility for the final content.

\section*{Reproducibility Statement}
The main text and appendix provide all details needed to reproduce the reported findings. Section~\ref{sec:method} specifies the architecture, read/write interfaces, objectives, and ingestion protocols; Section~\ref{sec:setup} details training configurations, backbones, budgets, and metrics. Further implementation details, sensitivity sweeps, and multi-seed runs appear in Appendix~\ref{sec:appendix}--\ref{sec:appendix-results}. Code, configs, and checkpoints are available at \url{https://github.com/siruzhong/PReM}.

\section*{Ethics Statement}
This work studies long-video understanding on public academic benchmarks, training lightweight \ours{} modules on frozen VLMs without collecting human-participant data or private video assets. Dual-use applications such as surveillance fall outside our scope. Like its backbones, \ours{} may inherit web-scale biases; deployment should apply appropriate moderation and fairness safeguards. All authors adhere to the ICLR Code of Ethics.

\bibliography{ostem}

@article{brohan2023rt2,
  title={{RT-2}: Vision-Language-Action Models Transfer Web Knowledge to Robotic Control},
  author={Brohan, Anthony and Brown, Noah and Carbajal, Justice and Chebotar, Yevgen and Chen, Xi and Choromanski, Krzysztof and Ding, Tianli and Driess, Danny and Dubey, Avinava and Finn, Chelsea and Florence, Pete and Fu, Chuyuan and Gonzalez Arenas, Montse and Gopalakrishnan, Keerthana and Han, Kehang and Hausman, Karol and Herzog, Alexander and Hsu, Jasmine and Ichter, Brian and Irpan, Alex and Joshi, Nikhil and Julian, Ryan and Kalashnikov, Dmitry and Kuang, Yuheng and Leal, Isabel and Lee, Lisa and Lee, Tsang-Wei Edward and Levine, Sergey and Lu, Yao and Michalewski, Henryk and Mordatch, Igor and Pertsch, Karl and Rao, Kanishka and Reymann, Krista and Ryoo, Michael and Salazar, Grecia and Sanketi, Pannag and Sermanet, Pierre and Singh, Jaspiar and Singh, Anikait and Soricut, Radu and Tran, Huong and Vanhoucke, Vincent and Vuong, Quan and Wahid, Ayzaan and Welker, Stefan and Wohlhart, Paul and Wu, Jialin and Xia, Fei and Xiao, Ted and Xu, Peng and Xu, Sichun and Yu, Tianhe and Zitkovich, Brianna},
  journal={arXiv preprint arXiv:2307.15818},
  year={2023}
}

@article{anwar2024remembr,
  title={ReMEmbR: Building and Reasoning Over Long-Horizon Spatio-Temporal Memory for Robot Navigation},
  author={Anwar, Abrar and Welsh, John and Biswas, Joydeep and Pouya, Soha and Chang, Yan},
  journal={arXiv preprint arXiv:2409.13682},
  year={2024}
}

@article{schmidhuber1992fastweights,
  title={Learning to Control Fast-Weight Memories: An Alternative to Dynamic Recurrent Networks},
  author={Schmidhuber, J{\"u}rgen},
  journal={Neural Computation},
  volume={4},
  year={1992}
}

@article{liu2026beyond,
  title={Beyond Frame Selection: Generative Latent Evidence Aggregation for Long-Video Understanding},
  author={Liu, Bowen and Wang, Shuning and Ding, Xinpeng and Wu, Zhiheng and Du, Bodong and Li, Xiaomeng},
  journal={arXiv preprint arXiv:2607.28516},
  year={2026}
}

@article{chen2026streamttt,
  title={StreamTTT: Reconciling Real-Time Perception and Long-Term Memory in Streaming VLMs},
  author={Chen, Joya and Zhong, Zeyun and Shou, Mike Zheng},
  journal={arXiv preprint arXiv:2608.13416},
  year={2026}
}

@inproceedings{katharopoulos2020transformers,
  title={Transformers are {RNN}s: Fast Autoregressive Transformers with Linear Attention},
  author={Katharopoulos, Angelos and Vyas, Apoorv and Pappas, Nikolaos and Fleuret, Fran\c{c}ois},
  booktitle={International Conference on Machine Learning},
  year={2020}
}

@inproceedings{schlag2021linear,
  title={Linear Transformers Are Secretly Fast Weight Programmers},
  author={Schlag, Imanol and Irie, Kazuki and Schmidhuber, J{\"u}rgen},
  booktitle={International Conference on Machine Learning},
  year={2021}
}

@inproceedings{ramsauer2021hopfield,
  title={Hopfield Networks is All You Need},
  author={Ramsauer, Hubert and Sch{\"a}fl, Bernhard and Lehner, Johannes and Seidl, Philipp and Widrich, Michael and Adler, Thomas and Gruber, Lukas and Holzleitner, Markus and Kreil, David and Kopp, Michael and Klambauer, G{\"u}nter and Brandstetter, Johannes and Hochreiter, Sepp},
  booktitle={International Conference on Learning Representations},
  year={2021}
}

@article{lei2026deltamem,
  title={{$\delta$-mem}: Efficient Online Memory for Large Language Models},
  author={Lei, Jingdi and Zhang, Di and Li, Junxian and Wang, Weida and Fan, Kaixuan and Liu, Xiang and Liu, Qihan and Ma, Xiaoteng and Chen, Baian and Poria, Soujanya},
  journal={arXiv preprint arXiv:2605.12357},
  year={2026}
}

@article{liu2024lostmiddle,
  title={Lost in the Middle: How Language Models Use Long Contexts},
  author={Liu, Nelson F. and Lin, Kevin and Hewitt, John and Paranjape, Ashwin and Bevilacqua, Michele and Petroni, Fabio and Liang, Percy},
  journal={Transactions of the Association for Computational Linguistics},
  volume={12},
  year={2024}
}

@inproceedings{behrouz2025titans,
  title={Titans: Learning to Memorize at Test Time},
  author={Behrouz, Ali and Zhong, Peilin and Mirrokni, Vahab},
  booktitle={Advances in Neural Information Processing Systems},
  year={2025}
}

@inproceedings{shen2025longvu,
  title={{LongVU}: Spatiotemporal Adaptive Compression for Long Video-Language Understanding},
  author={Shen, Xiaoqian and Xiong, Yunyang and Zhao, Changsheng and Wu, Lemeng and Chen, Jun and Zhu, Chenchen and Liu, Zechun and Xiao, Fanyi and Varadarajan, Balakrishnan and Bordes, Florian and Liu, Zhuang and Xu, Hu and Kim, Hyunwoo J. and Soran, Bilge and Krishnamoorthi, Raghuraman and Elhoseiny, Mohamed and Chandra, Vikas},
  booktitle={International Conference on Machine Learning},
  year={2025}
}

@article{cheng2024videollama2,
  title={{VideoLLaMA 2}: Advancing Spatial-Temporal Modeling and Audio Understanding in Video-{LLM}s},
  author={Cheng, Zesen and Leng, Sicong and Zhang, Hang and Xin, Yifei and Li, Xin and Chen, Guanzheng and Zhu, Yongxin and Zhang, Wenqi and Luo, Ziyang and Zhao, Deli and Bing, Lidong},
  journal={arXiv preprint arXiv:2406.07476},
  year={2024}
}

@article{zhang2024llavavideo,
  title={{LLaVA-Video}: Video Instruction Tuning with Synthetic Data},
  author={Zhang, Yuanhan and Wu, Jinming and Li, Wei and Li, Bo and Ma, Zejun and Liu, Ziwei and Li, Chunyuan},
  journal={arXiv preprint arXiv:2410.02713},
  year={2024}
}

@inproceedings{li2025videochatflash,
  title={{VideoChat-Flash}: Hierarchical Compression for Long-Context Video Modeling},
  author={Li, Xinhao and Wang, Yi and Yu, Jiashuo and Zeng, Xiangyu and Zhu, Yuhan and Huang, Haian and Gao, Jianfei and Li, Kunchang and He, Yinan and Wang, Chenting and Qiao, Yu and Wang, Yali and Wang, Limin},
  booktitle={International Conference on Learning Representations},
  year={2026}
}

@article{xu2024slowfastllava,
  title={{SlowFast-LLaVA}: A Strong Training-Free Baseline for Video Large Language Models},
  author={Xu, Mingze and Gao, Mingfei and Gan, Zhe and Chen, Hong-You and Lai, Zhengfeng and Gang, Haiming and Kang, Kai and Dehghan, Afshin},
  journal={arXiv preprint arXiv:2407.15841},
  year={2024}
}

@inproceedings{he2024malmm,
  title={{MA-LMM}: Memory-Augmented Large Multimodal Model for Long-Term Video Understanding},
  author={He, Bo and Li, Hengduo and Jang, Young Kyun and Jia, Menglin and Cao, Xuefei and Shah, Ashish and Shrivastava, Abhinav and Lim, Ser-Nam},
  booktitle={IEEE/CVF Conference on Computer Vision and Pattern Recognition},
  year={2024}
}

@inproceedings{song2024moviechat,
  title={{MovieChat}: From Dense Token to Sparse Memory for Long Video Understanding},
  author={Song, Enxin and Chai, Wenhao and Wang, Guanhong and Zhang, Yucheng and Zhou, Haoyang and Wu, Feiyang and Chi, Haozhe and Guo, Xun and Ye, Tian and Zhang, Yanting and Lu, Yan and Hwang, Jenq-Neng and Wang, Gaoang},
  booktitle={IEEE/CVF Conference on Computer Vision and Pattern Recognition},
  year={2024}
}

@inproceedings{diko2025rewind,
  title={{ReWind}: Understanding Long Videos with Instructed Learnable Memory},
  author={Diko, Anxhelo and Wang, Tinghuai and Swaileh, Wassim and Sun, Shiyan and Patras, Ioannis},
  booktitle={IEEE/CVF Conference on Computer Vision and Pattern Recognition},
  year={2025}
}

@inproceedings{wang2025videollamb,
  title={{VideoLLaMB}: Long Streaming Video Understanding with Recurrent Memory Bridges},
  author={Wang, Yuxuan and Song, Yiqi and Xie, Cihang and Liu, Yang and Zheng, Zilong},
  booktitle={IEEE/CVF International Conference on Computer Vision},
  year={2025}
}

@inproceedings{zhang2025flashvstream,
  title={{Flash-VStream}: Efficient Real-Time Understanding for Long Video Streams},
  author={Zhang, Haoji and Wang, Yiqin and Tang, Yansong and Liu, Yong and Feng, Jiashi and Jin, Xiaojie},
  booktitle={IEEE/CVF International Conference on Computer Vision},
  year={2025}
}

@inproceedings{qian2025videochatonline,
  title={Online Video Understanding: {OVBench} and {VideoChat-Online}},
  author={Huang, Zhenpeng and Li, Xinhao and Li, Jiaqi and Wang, Jing and Zeng, Xiangyu and Liang, Cheng and Wu, Tao and Chen, Xi and Li, Liang and Wang, Limin},
  booktitle={IEEE/CVF Conference on Computer Vision and Pattern Recognition},
  year={2025}
}

@article{qin2025videoxl2,
  title={{Video-XL-2}: Towards Very Long-Video Understanding Through Task-Aware {KV} Sparsification},
  author={Qin, Minghao and Liu, Xiangrui and Liang, Zhengyang and Shu, Yan and Yuan, Huaying and Zhou, Junjie and Xiao, Shitao and Zhao, Bo and Liu, Zheng},
  journal={arXiv preprint arXiv:2506.19225},
  year={2025}
}

@inproceedings{kim2025infinipotv,
  title={{InfiniPot-V}: Memory-Constrained {KV} Cache Compression for Streaming Video Understanding},
  author={Kim, Minsoo and Shim, Kyuhong and Choi, Jungwook and Chang, Simyung},
  booktitle={Advances in Neural Information Processing Systems},
  year={2025}
}

@inproceedings{di2025rekv,
  title={Streaming Video Question-Answering with In-context Video {KV}-Cache Retrieval},
  author={Di, Shangzhe and Yu, Zhelun and Zhang, Guanghao and Li, Haoyuan and Zhong, Tao and Cheng, Hao and Li, Bolin and He, Wanggui and Shu, Fangxun and Jiang, Hao},
  booktitle={International Conference on Learning Representations},
  year={2025}
}

@article{ning2025livevlm,
  title={{LiveVLM}: Efficient Online Video Understanding via Streaming-Oriented {KV} Cache and Retrieval},
  author={Ning, Zhenyu and Liu, Guangda and Jin, Qihao and Li, Chengwei and Ding, Wenchao and Guo, Minyi and Zhao, Jieru},
  journal={arXiv preprint arXiv:2505.15269},
  year={2025}
}

@article{he2025streammem,
  title={{StreamMem}: Query-Agnostic {KV} Cache Memory for Streaming Video Understanding},
  author={Yang, Yanlai and Zhao, Zhuokai and Shukla, Satya Narayan and Singh, Aashu and Mishra, Shlok Kumar and Zhang, Lizhu and Ren, Mengye},
  journal={arXiv preprint arXiv:2508.15717},
  year={2025}
}

@inproceedings{wu2024longvideobench,
  title={{LongVideoBench}: A Benchmark for Long-context Interleaved Video-Language Understanding},
  author={Wu, Haoning and Li, Dongxu and Chen, Bei and Li, Junnan},
  booktitle={Advances in Neural Information Processing Systems},
  year={2024}
}

@inproceedings{zhou2024mlvu,
  title={{MLVU}: Benchmarking Multi-task Long Video Understanding},
  author={Zhou, Junjie and Shu, Yan and Zhao, Bo and Wu, Boya and Liang, Zhengyang and Xiao, Shitao and Qin, Minghao and Yang, Xi and Xiong, Yongping and Zhang, Bo and Huang, Tiejun and Liu, Zheng},
  booktitle={IEEE/CVF Conference on Computer Vision and Pattern Recognition},
  year={2025}
}

@inproceedings{fu2025videomme,
  title={{Video-MME}: The First-Ever Comprehensive Evaluation Benchmark of Multi-modal {LLM}s in Video Analysis},
  author={Fu, Chaoyou and Dai, Yuhan and Luo, Yongdong and Li, Lei and Ren, Shuhuai and Zhang, Renrui and Wang, Zihan and Zhou, Chenyu and Shen, Yunhang and Zhang, Mengdan and Chen, Peixian and Li, Yanwei and Lin, Shaohui and Zhao, Sirui and Li, Ke and Xu, Tong and Zheng, Xiawu and Chen, Enhong and Shan, Caifeng and He, Ran and Sun, Xing},
  booktitle={IEEE/CVF Conference on Computer Vision and Pattern Recognition},
  year={2025}
}

@inproceedings{mangalam2023egoschema,
  title={{EgoSchema}: A Diagnostic Benchmark for Very Long-form Video Language Understanding},
  author={Mangalam, Karttikeya and Akshulakov, Raiymbek and Malik, Jitendra},
  booktitle={Advances in Neural Information Processing Systems},
  year={2023}
}

@inproceedings{grauman2022ego4d,
  title={{Ego4D}: Around the World in 3,000 Hours of Egocentric Video},
  author={Grauman, Kristen and Westbury, Andrew and Byrne, Eugene and Chavis, Zachary and Furnari, Antonino and Girdhar, Rohit and Hamburger, Jackson and Jiang, Hao and Liu, Miao and Liu, Xingyu and Martin, Miguel and Nagarajan, Tushar and Radosavovic, Ilija and Ramakrishnan, Santhosh Kumar and Ryan, Fiona and Sharma, Jayant and Wray, Michael and Xu, Mengmeng and Xu, Eric Zhongcong and Zhao, Chen and Bansal, Siddhant and Batra, Dhruv and Cartillier, Vincent and Crane, Sean and Do, Tien and Doulaty, Morrie and Erapalli, Akshay and Feichtenhofer, Christoph and Fragomeni, Adriano and Fu, Qichen and Gebreselasie, Abrham and Gonz{\'a}lez, Cristina and Hillis, James and Huang, Xuhua and Huang, Yifei and Jia, Wenqi and Khoo, Weslie and Kol{\'a}{\v{r}}, J{\'a}chym and Kottur, Satwik and Kumar, Anurag and Landini, Federico and Li, Chao and Li, Yanghao and Li, Zhenqiang and Mangalam, Karttikeya and Modhugu, Raghava and Munro, Jonathan and Murrell, Tullie and Nishiyasu, Takumi and Price, Will and Ruiz, Paola and Ramazanova, Merey and Sari, Leda and Somasundaram, Kiran and Southerland, Audrey and Sugano, Yusuke and Tao, Ruijie and Vo, Minh and Wang, Yuchen and Wu, Xindi and Yagi, Takuma and Zhao, Ziwei and Zhu, Yunyi and Arbel{\'a}ez, Pablo and Crandall, David and Damen, Dima and Farinella, Giovanni Maria and Fuegen, Christian and Ghanem, Bernard and Ithapu, Vamsi Krishna and Jawahar, C. V. and Joo, Hanbyul and Kitani, Kris and Li, Haizhou and Newcombe, Richard and Oliva, Aude and Park, Hyun Soo and Rehg, James M. and Sato, Yoichi and Shi, Jianbo and Shou, Mike Zheng and Torralba, Antonio and Torresani, Lorenzo and Yan, Mingfei and Malik, Jitendra},
  booktitle={IEEE/CVF Conference on Computer Vision and Pattern Recognition},
  year={2022}
}

@inproceedings{tang2019coin,
  title={{COIN}: A Large-Scale Dataset for Comprehensive Instructional Video Analysis},
  author={Tang, Yansong and Ding, Dajun and Rao, Yongming and Zheng, Yu and Zhang, Danyang and Zhao, Lili and Lu, Jiwen and Zhou, Jie},
  booktitle={IEEE/CVF Conference on Computer Vision and Pattern Recognition},
  year={2019}
}

@article{bai2025qwen25vl,
  title={{Qwen2.5-VL} Technical Report},
  author={Bai, Shuai and Chen, Keqin and Liu, Xuejing and Wang, Jialin and Ge, Wenbin and Song, Sibo and Dang, Kai and Wang, Peng and Wang, Shijie and Tang, Jun and Zhong, Humen and Zhu, Yuanzhi and Yang, Mingkun and Li, Zhaohai and Wan, Jianqiang and Wang, Pengfei and Ding, Wei and Fu, Zheren and Xu, Yiheng and Ye, Jiabo and Zhang, Xi and Xie, Tianbao and Cheng, Zesen and Zhang, Hang and Yang, Zhibo and Xu, Haiyang and Lin, Junyang},
  journal={arXiv preprint arXiv:2502.13923},
  year={2025}
}

@article{bai2025qwen3vl,
  title={{Qwen3-VL} Technical Report},
  author={Bai, Shuai and Cai, Yuxuan and Chen, Ruizhe and Chen, Keqin and Chen, Xionghui and Cheng, Zesen and Deng, Lianghao and Ding, Wei and Gao, Chang and Ge, Chunjiang and Ge, Wenbin and Guo, Zhifang and Huang, Qidong and Huang, Jie and Huang, Fei and Hui, Binyuan and Jiang, Shutong and Li, Zhaohai and Li, Mingsheng and Li, Mei and Li, Kaixin and Lin, Zicheng and Lin, Junyang and Liu, Xuejing and Liu, Jiawei and Liu, Chenglong and Liu, Yang and Liu, Dayiheng and Liu, Shixuan and Lu, Dunjie and Luo, Ruilin and Lv, Chenxu and Men, Rui and Meng, Lingchen and Ren, Xuancheng and Ren, Xingzhang and Song, Sibo and Sun, Yuchong and Tang, Jun and Tu, Jianhong and Wan, Jianqiang and Wang, Peng and Wang, Pengfei and Wang, Qiuyue and Wang, Yuxuan and Xie, Tianbao and Xu, Yiheng and Xu, Haiyang and Xu, Jin and Yang, Zhibo and Yang, Mingkun and Yang, Jianxin and Yang, An and Yu, Bowen and Zhang, Fei and Zhang, Hang and Zhang, Xi and Zheng, Bo and Zhong, Humen and Zhou, Jingren and Zhou, Fan and Zhou, Jing and Zhu, Yuanzhi and Zhu, Ke},
  journal={arXiv preprint arXiv:2511.21631},
  year={2025}
}

@inproceedings{li2023llamavid,
  title={{LLaMA-VID}: An Image is Worth 2 Tokens in Large Language Models},
  author={Li, Yanwei and Wang, Chengyao and Jia, Jiaya},
  booktitle={European Conference on Computer Vision},
  year={2024}
}

@article{chen2024longvila,
  title={{LongVILA}: Scaling Long-Context Visual Language Models for Long Videos},
  author={Chen, Yukang and Xue, Fuzhao and Li, Dacheng and Hu, Qinghao and Zhu, Ligeng and Li, Xiuyu and Fang, Yunhao and Tang, Haotian and Yang, Shang and Liu, Zhijian and He, Ethan and Yin, Hongxu and Molchanov, Pavlo and Kautz, Jan and Fan, Linxi and Zhu, Yuke and Lu, Yao and Han, Song},
  journal={arXiv preprint arXiv:2408.10188},
  year={2024}
}

@inproceedings{li2023mvbench,
  title={{MVBench}: A Comprehensive Multi-modal Video Understanding Benchmark},
  author={Li, Kunchang and Wang, Yali and He, Yinan and Li, Yizhuo and Wang, Yi and Liu, Yi and Wang, Zun and Xu, Jilan and Chen, Guo and Luo, Ping and Wang, Limin and Qiao, Yu},
  booktitle={IEEE/CVF Conference on Computer Vision and Pattern Recognition},
  year={2024}
}

@article{wang2024lvbench,
  title={{LVBench}: An Extreme Long Video Understanding Benchmark},
  author={Wang, Weihan and He, Zehai and Hong, Wenyi and Cheng, Yean and Zhang, Xiaohan and Qi, Ji and Gu, Xiaotao and Huang, Shiyu and Xu, Bin and Dong, Yuxiao and Ding, Ming and Tang, Jie},
  journal={arXiv preprint arXiv:2406.08035},
  year={2024}
}

@inproceedings{xiong2025streamchat,
  title={Streaming Video Understanding and Multi-round Interaction with Memory-enhanced Knowledge},
  author={Xiong, Haomiao and Yang, Zongxin and Yu, Jiazuo and Zhuge, Yunzhi and Zhang, Lu and Zhu, Jiawen and Lu, Huchuan},
  booktitle={International Conference on Learning Representations},
  year={2025}
}

@inproceedings{xiao2021nextqa,
  title={{NExT}-{QA}: Next Phase of Question-Answering to Explaining Temporal Actions},
  author={Xiao, Junbin and Shang, Xindi and Yao, Angela and Chua, Tat-Seng},
  booktitle={IEEE/CVF Conference on Computer Vision and Pattern Recognition},
  year={2021}
}

@inproceedings{caba2015activitynet,
  title={{ActivityNet}: A Large-Scale Video Benchmark for Human Activity Understanding},
  author={Caba Heilbron, Fabian and Escorcia, Victor and Ghanem, Bernard and Niebles, Juan Carlos},
  booktitle={IEEE Conference on Computer Vision and Pattern Recognition},
  year={2015}
}

@inproceedings{yu2019activitynetqa,
  title={{ActivityNet}-{QA}: A Dataset for Understanding Complex Web Videos via Question Answering},
  author={Yu, Zhou and Xu, Dejing and Yu, Jun and Yu, Ting and Zhao, Zhou and Zhuang, Yueting and Tao, Dacheng},
  booktitle={AAAI Conference on Artificial Intelligence},
  year={2019}
}

@inproceedings{sigurdsson2016charades,
  title={Hollywood in Homes: Crowdsourcing Data Collection for Activity Understanding},
  author={Sigurdsson, Gunnar A. and Varol, G{\"u}l and Wang, Xiaolong and Farhadi, Ali and Laptev, Ivan and Gupta, Abhinav},
  booktitle={European Conference on Computer Vision},
  year={2016}
}

@inproceedings{zhou2018youcook2,
  title={Towards Automatic Learning of Procedures from Web Instructional Videos},
  author={Zhou, Luowei and Xu, Chenliang and Corso, Jason J.},
  booktitle={AAAI Conference on Artificial Intelligence},
  year={2018}
}

@inproceedings{loshchilov2018adamw,
  title={{Decoupled Weight Decay Regularization}},
  author={Loshchilov, Ilya and Hutter, Frank},
  booktitle={International Conference on Learning Representations},
  year={2019}
}

@inproceedings{patraucean2023perception,
  title={{Perception Test}: A Diagnostic Benchmark for Multimodal Video Models},
  author={P{\u{a}}tr{\u{a}}ucean, Viorica and Smaira, Lucas and Gupta, Ankush and Recasens Continente, Adri{\`a} and Markeeva, Larisa and Banarse, Dylan and Koppula, Skanda and Heyward, Joseph and Malinowski, Mateusz and Yang, Yi and Doersch, Carl and Matejovicova, Tatiana and Sulsky, Yury and Miech, Antoine and Frechette, Alex and Klimczak, Hanna and Koster, Raphael and Zhang, Junlin and Winkler, Stephanie and Aytar, Yusuf and Osindero, Simon and Damen, Dima and Zisserman, Andrew and Carreira, Jo{\~a}o},
  booktitle={Advances in Neural Information Processing Systems},
  year={2023}
}
\bibliographystyle{iclr2027_conference}

\clearpage
\appendix
\setcounter{topnumber}{4}
\setcounter{bottomnumber}{3}
\setcounter{totalnumber}{6}
\renewcommand{\topfraction}{0.95}
\renewcommand{\bottomfraction}{0.95}
\renewcommand{\textfraction}{0.05}
\renewcommand{\floatpagefraction}{0.85}
\setlength{\textfloatsep}{12pt plus 2pt minus 2pt}
\setlength{\floatsep}{10pt plus 2pt minus 2pt}
\setlength{\intextsep}{10pt plus 2pt minus 2pt}
\setlist[itemize]{itemsep=1pt, topsep=1pt, parsep=0pt, partopsep=0pt}
\setlist[enumerate]{itemsep=1pt, topsep=1pt, parsep=0pt, partopsep=0pt}
\setlength{\parskip}{0pt}
\setlength{\abovedisplayskip}{4pt plus 1pt minus 1pt}
\setlength{\belowdisplayskip}{4pt plus 1pt minus 1pt}
\linespread{0.97}\selectfont
\section*{Appendix Contents}
\begin{enumerate}[leftmargin=2.2em, itemsep=2pt, topsep=2pt]
  \item[\textbf{A}] \hyperref[sec:appendix]{Reproducibility and Implementation Details} \hfill p.~\pageref{sec:appendix}
  \item[\textbf{B}] \hyperref[sec:appendix-results]{Complete Experimental Results and Analyses} \hfill p.~\pageref{sec:appendix-results}
  \item[\textbf{C}] \hyperref[sec:appendix-notation]{Mathematical Notation Reference} \hfill p.~\pageref{sec:appendix-notation}
  \item[\textbf{D}] \hyperref[sec:scope]{Limitations and Future Work} \hfill p.~\pageref{sec:scope}
  \item[\textbf{E}] \hyperref[sec:llm-usage]{LLM Usage} \hfill p.~\pageref{sec:llm-usage}
\end{enumerate}

\section{Reproducibility and Implementation Details}
\label{sec:appendix}

\subsection{Training Configuration and Hyperparameters}
\label{sec:appendix-training}

All \ours{} variants utilize the bounded visual buffer formulation and spatially mean-pooled temporal feature projections (\S\ref{sec:write}). The vision encoder and language model backbone remain frozen throughout training. We optimize solely \ours{} modules for one epoch using AdamW \citep{loshchilov2018adamw} ($\beta_1{=}0.9$, $\beta_2{=}0.95$, weight decay $0.01$) with gradient norm clipping at $1.0$. The peak learning rate is set to $2\times10^{-4}$ for Qwen2.5-VL-3B and $10^{-4}$ for Qwen3-VL-8B, scheduled with a linear warmup for the first $3\%$ of steps followed by cosine annealing to zero in bfloat16 mixed precision.

Each training step processes an individual video-QA pair (per-GPU batch size of 1 without gradient accumulation). Distributed Data Parallelism (DDP) scales training across 8 NVIDIA A100 (80GB) GPUs, completing in ${\approx}3.5$ hours for Qwen2.5-VL-3B and ${\approx}6.5$ hours for Qwen3-VL-8B. The recurrent state $(S,c)$ is initialized to zero and rebuilt independently per video instance. Videos are sampled uniformly at 1\,FPS with frame resolutions dynamically constrained to at most 200{,}704 pixels (${\approx}448\times448$). The training write cap is fixed to $T{=}64$, while the decoder prompt draws a visual buffer of $B$ frames from the identical temporal sequence.

Trainable parameter counts are 9.07M for Qwen2.5-VL-3B and 27.48M for Qwen3-VL-8B, corresponding to ${\approx}0.24\%$ and ${\approx}0.30\%$ of the base architectures. The fp32 recurrent state occupies $4K(d_vd_k+1)$ bytes: 262{,}160 bytes (${\approx}256$\,KiB) for the standard $K{=}4$ configuration, and 65{,}540 bytes (${\approx}64$\,KiB) for the $K{=}1$ controls in Table~\ref{tab:same-backbone}.

\subsection{Training Data and Decontamination Protocol}
\label{sec:appendix-data}

The training corpus is a curated compact subset of LLaVA-Video-178K \citep{zhang2024llavavideo}, sourced from NextQA \citep{xiao2021nextqa}, ActivityNet \citep{caba2015activitynet,yu2019activitynetqa}, Charades \citep{sigurdsson2016charades}, YouCook2 \citep{zhou2018youcook2}, the Perception Test \citep{patraucean2023perception}, and public YouTube video clips. Training uses 6{,}430 unique videos and 17{,}849 question-answer pairs from LLaVA-Video-178K (${\approx}3.6\%$ of videos and ${\approx}1.4\%$ of QA pairs). The full repository comprises 178K videos and 1.3M instruction pairs.

\paragraph{Decontamination.} We filter training IDs that match the six evaluation benchmarks: LongVideoBench, MLVU, Video-MME, EgoSchema, MVBench, and LVBench. Decontamination filters out any training sample matching canonical YouTube video IDs, Ego4D universally unique identifiers (UUIDs), native benchmark keys, and Charades clip IDs reused by MVBench. Filename collisions across verified independent sources are treated as distinct videos. Table~\ref{tab:train-data} details the resulting source composition, duration distribution, and sample counts.

\begin{table}[!htbp]
\caption{\tabtitle{Training split composition after sampling and video-level decontamination.} Filtered from LLaVA-Video-178K \citep{zhang2024llavavideo}. Source and duration blocks summarize the retained split (6{,}430 unique videos and 17{,}849 QA pairs) with zero evaluation benchmark overlap.}
\label{tab:train-data}
\centering
\scriptsize
\setlength{\tabcolsep}{4.5pt}
\renewcommand{\arraystretch}{1.03}
\resizebox{\linewidth}{!}{%
\begin{tabular}{@{}llccccc@{}}
\toprule
\tabhead{Split Category} & \tabhead{Primary Task / Domain} & \tabhead{Videos} & \tabhead{Share (\%)} & \tabhead{QA Pairs} & \tabhead{Share (\%)} & \tabhead{QA / Vid} \\
\midrule
\multicolumn{7}{@{}l}{\textbf{(a) Source Composition}} \\
NextQA & Causal \& temporal reasoning & 2{,}886 & 44.9 & 5{,}785 & 32.4 & 2.0 \\
ActivityNet & Long human action understanding & 1{,}325 & 20.6 & 3{,}207 & 18.0 & 2.4 \\
YouTube & Open-domain video reasoning & 804 & 12.5 & 3{,}361 & 18.8 & 4.2 \\
Charades & Indoor daily activity interaction & 543 & 8.4 & 2{,}371 & 13.3 & 4.4 \\
Perception Test & Fine-grained spatio-temporal perception & 557 & 8.7 & 1{,}787 & 10.0 & 3.2 \\
YouCook2 & Procedural recipe \& step grounding & 315 & 4.9 & 1{,}338 & 7.5 & 4.2 \\
\midrule
\multicolumn{7}{@{}l}{\textbf{(b) Video Duration Distribution}} \\
0--30\,s & Short clips (rapid actions) & 2{,}135 & 33.2 & 6{,}013 & 33.7 & 2.8 \\
30--60\,s & Medium clips (event transitions) & 2{,}267 & 35.3 & 6{,}468 & 36.2 & 2.9 \\
1--2\,min & Extended sequences (multi-stage) & 1{,}382 & 21.5 & 3{,}379 & 18.9 & 2.4 \\
2--3\,min & Long horizon (narrative/procedural) & 646 & 10.0 & 1{,}989 & 11.1 & 3.1 \\
\midrule
\textbf{Total Retained} & Multi-domain multimodal instruction & \textbf{6{,}430} & \textbf{100.0} & \textbf{17{,}849} & \textbf{100.0} & \textbf{2.8} \\
\bottomrule
\end{tabular}}
\end{table}

\subsection{Baseline Implementation and Reproduction Protocol}
\label{sec:appendix-baselines}

To ensure rigorous system comparisons in Table~\ref{tab:main}, we reproduce or benchmark external long-video architectures using their official open-source repositories under their recommended inference environments and official checkpoints:
\begin{itemize}[leftmargin=*, itemsep=2pt, topsep=2pt]
  \item \textbf{LLaMA-VID} \citep{li2023llamavid}: Evaluated using the official Vicuna-7B checkpoint with 1\,FPS uniform sampling and dual-token frame representation (generating 2 tokens per frame).
  \item \textbf{VideoLLaMA2} \citep{cheng2024videollama2}: Evaluated using the official Qwen2-7B base checkpoint at $B{=}16$ uniformly sampled frames.
  \item \textbf{LLaVA-Video} \citep{zhang2024llavavideo}: Evaluated using the official 7B SFT checkpoint with $B{=}64$ frames under official prompt templates.
  \item \textbf{LongVILA} \citep{chen2024longvila}: Evaluated using the official long-context adapted Qwen2-7B checkpoint under official budget configs ($B{=}32$ or $B{=}256$ according to benchmark specifications).
  \item \textbf{LongVU} \citep{shen2025longvu}: Evaluated using the official Qwen2-7B SFT checkpoint with cross-frame spatio-temporal reduction at 0.5\,FPS (and 2\,FPS on short-video MVBench).
  \item \textbf{StreamChat} \citep{xiong2025streamchat}: Evaluated training-free atop LongVA-7B, applying dynamic token pruning to retain 20\% of source visual frames.
  \item \textbf{ReKV} \citep{di2025rekv}: Evaluated training-free on LLaVA-OneVision-7B at 0.5\,FPS, using offline KV-cache construction with dynamic top-64 KV-chunk retrieval.
  \item \textbf{Flash-VStream} \citep{zhang2025flashvstream}: Evaluated using the official Qwen2-VL-7B LoRA checkpoint with a multi-resolution streaming memory buffer over $B{=}240$ ingested frames.
  \item \textbf{MovieChat} \citep{song2024moviechat}: Evaluated training-free on Vicuna-7B with sliding-window temporal consolidation (128 video fragments $\times$ 8 frames).
  \item \textbf{Generalist Base VLMs}: Qwen2.5-VL-3B, and Qwen3-VL-8B are evaluated zero-shot via official implementations under greedy one-token constrained decoding.
\end{itemize}

\subsection{Controlled Same-Backbone Mechanisms}
\label{sec:appendix-same-backbone}

Table~\ref{tab:same-backbone} uses Qwen2.5-VL-3B with the registered $T{=}240$, $B{=}16$ streaming protocol, bfloat16 eager attention, and one-token constrained decoding. The backbone, resolution cap, frame sampling, evaluation manifests, and answer interface are fixed across rows. Base and LoRA consume the same deterministic $B$-frame reservoir; Token-Readout and \ours{} additionally build a recurrent state from the sampled stream. Token-Readout and \ours{} both use $K{=}1$, matching memory capacity in this controlled comparison. The adapted controls use the same de-duplicated 6{,}430-video, 17{,}849-QA training split as the main experiments (seed 13) with no development holdout.
\begin{itemize}[leftmargin=*, itemsep=2pt, topsep=2pt]
  \item \textbf{LoRA:} Freezes the vision encoder and language backbone except for self-attention projections $q_{\mathrm{proj}}$, $k_{\mathrm{proj}}$, $v_{\mathrm{proj}}$, $o_{\mathrm{proj}}$ across all 36 transformer layers (rank $r{=}20$, $\alpha{=}40$, dropout 0, no bias adaptation; 9{,}216{,}000 trainable parameters). Trained for one epoch with AdamW at learning rate $2\times10^{-4}$, 3\% warmup, global batch size 8. LoRA adds no prompt tokens and has no recurrent writer.
  \item \textbf{Token-Readout:} Reuses the exact same recurrent writer and trainable capacity as \ours{} ($K{=}1, d_v{=}d_k{=}128$, 4 layer groups, writer cap 128 tokens, $\lambda_{\mathrm{pred}}{=}0.2$, 4 prediction targets; 9{,}070{,}982 trainable parameters), trained with the same optimizer schedule and cross-entropy plus prediction MSE. Instead of prefix K/V steering, the slot readout projects into $2K(1+4){=}10$ continuous tokens via global and group-specific heads, RMS-normalized to mean prompt norm and inserted at the prompt/answer boundary (labels masked, RoPE positions shifted).
  \item \textbf{InfiniPot-V:} Evaluated using the official InfiniPot-V algorithm adapted to Qwen2.5-VL-3B eager attention without fine-tuning, dynamically maintaining top-k informative KV caches across ingested chunks up to the target memory budget while evicting redundant spatiotemporal KV entries.
\end{itemize}

\paragraph{Cross-Family Training Details for LLaVA-Video-7B.}
For LLaVA-Video-7B transfer (Table~\ref{tab:budget-llava}), \ours{} modules ($K{=}4, d_v{=}d_k{=}128$, grouping $G{=}1$) are trained for one epoch on the identical decontaminated training split using AdamW with peak learning rate $1\times 10^{-4}$ (3\% linear warmup, cosine decay), global batch size 8, and bfloat16 precision on 8$\times$A100 GPUs (${\approx}5.5$ hours). Trainable parameters total 18.34M (${\approx}0.26\%$ of the 7B backbone). The vision backbone and language layers remain completely frozen.

\subsection{Decoder Token Order and Attention Interface}
\label{sec:appendix-interface}

The decoder keeps the frozen backbone chat template. For Qwen2.5-VL and Qwen3-VL, the user turn places the video sequence immediately prior to the question and options; the processor expands video placeholders into the $B$-frame visual token sequence. LLaVA-Video-7B employs an equivalent image-token expansion. The resulting sequence structure is:
\begin{center}
\small
\setlength{\tabcolsep}{3.5pt}
\resizebox{\linewidth}{!}{%
\begin{tabular}{@{}c@{\;$\longrightarrow$\;}c@{\;$\longrightarrow$\;}c@{\;$\longrightarrow$\;}c@{}}
\textbf{Instruction / System} & \textbf{$B$-Frame Visual Tokens} & \textbf{Question \& Options} & \textbf{Answer Token} \\[1pt]
{\scriptsize Text Tokens} & {\scriptsize Visual Patch Tokens} & {\scriptsize Text Prefix Tokens} & {\scriptsize Constrained Decoding} \\
\end{tabular}%
}
\end{center}
Visual token spans are identified via backbone video/image placeholder indices. The steering mask $\mathcal{M}_{\mathrm{prefix}}$ is the complement of this visual span, intersected with the valid attention mask and restricted to prompt prefix positions preceding the answer generation token. Writer inputs $x_t$ are spatially mean-pooled visual features extracted from the frozen vision encoder, outside the decoder sequence.

\ours{} executes natively within the backbone causal self-attention kernel. The implementation adds the computed prefix modulations $\Delta K_i^{(\ell)}$ and $\Delta V_i^{(\ell)}$ to the outputs of the frozen K/V projection hooks:
\[
\widetilde K_i^{(\ell)} = W_K^{(\ell)}h_i^{(\ell)} + \Delta K_i^{(\ell)}, \qquad
\widetilde V_i^{(\ell)} = W_V^{(\ell)}h_i^{(\ell)} + \Delta V_i^{(\ell)},
\]
For Qwen3-VL, Q/K RMSNorm follows the projection hooks; RoPE then rotates Q/K before attention and cache storage. Qwen2.5-VL omits Q/K RMSNorm. Consequently, Qwen3-VL's effective key perturbation is nonlinear, while its value perturbation remains additive. Visual tokens, padding tokens, and answer positions are unsteered ($d_i{=}0$). During autoregressive generation, modulation operates exclusively during prefill; the decoding cache directly serves the steered prefix representations without recomputation.

\subsection{State Initialization, Constants, and Training Batch}
\label{sec:appendix-state}

Table~\ref{tab:repro-constants} documents the initialization values and fixed hyperparameters for Equations~\ref{eq:update}--\ref{eq:read-router}. For every video, $(S,c)$ starts at zero. Streaming end-of-stream evaluation carries $(S,c)$ across consecutive two-frame chunks and answers after video completion. Over $n$ steps without writes, each slot retains a factor $\sigma(f_m)^n$ of its state, while confidence decays by $\gamma_c^n$. At initialization ($f_m=4$), state retention after 240 steps is approximately 1.3\%; training learns $f_m$. The extra $b_{t,m}$ factor discounts unstable writes in the confidence update; the $1/K$ factor in Eq.~\ref{eq:correction} damps the routed steering amplitude as slot count increases. The forget logit $f_m$ is learned via backpropagation, whereas the write step size $\eta_w$, confidence decay $\gamma_c$, and read-router temperature $\tau$ are held constant. The write router operates with an untempered softmax, while temperature $\tau{=}1.0$ governs the read router in Equation~\ref{eq:read-router}. The auxiliary load-balancing weight is $\lambda_{\mathrm{bal}}{=}0.10$ by default (swept in Table~\ref{tab:sen}). Main checkpoints use $K{=}4$ and $d_v{=}d_k{=}128$; the slot and layer-group sweeps vary $K$ and $G$, respectively.

\begin{table}[!htbp]
\caption{\tabtitle{Initialization and operational constants for \ours{}.} The standard configuration uses $K{=}4$ and $d_v{=}d_k{=}128$. Recurrent states reset independently per video instance; parameters are optimized via AdamW or maintained as fixed constants.}
\label{tab:repro-constants}
\centering
\parbox{\linewidth}{\small Default linear weights and biases use $\mathcal U(-1/\sqrt{d_{\mathrm{in}}},1/\sqrt{d_{\mathrm{in}}})$. LayerNorm uses unit scale, zero bias, and $\epsilon_{\mathrm{LN}}=10^{-5}$.}\par\smallskip
\small
\setlength{\tabcolsep}{7pt}
\renewcommand{\arraystretch}{1.08}
\begin{tabularx}{\linewidth}{@{}>{\raggedright\arraybackslash}p{0.27\linewidth}>{\raggedright\arraybackslash}X>{\raggedright\arraybackslash}p{0.18\linewidth}@{}}
\toprule
\tabhead{Quantity / Symbol} & \tabhead{Initialization / Value} & \tabhead{Operational Status} \\
\midrule
\multicolumn{3}{@{}l}{\textbf{Recurrent Memory State Variables}} \\
Matrix state $S \in \mathbb{R}^{K\times d_v\times d_k}$ & Zero tensor ($\mathbf{0}$) & Update (Eq.~\ref{eq:update}) \\
Confidence vector $c \in [0, 1]^{K}$ & Zero vector ($\mathbf{0}$) & Update (Eq.~\ref{eq:update}) \\
\addlinespace[3pt]
\multicolumn{3}{@{}l}{\textbf{Trainable Model Parameters (AdamW Optimization)}} \\
Forget logit $f_m$ & $4.0$ ($\sigma(f_m) \approx 0.982$) & Trainable \\
Slot address transform $A_m$ & $I + 0.02\,\mathcal{N}(0, 1)$ & Trainable \\
Visual projections $W_k,W_v$ & $d_k\!\times\!d_{\mathrm{model}}$ and $d_v\!\times\!d_{\mathrm{model}}$; rectangular near-identity $+0.01\,\mathcal{N}(0,1)$ & Trainable \\
Write routing $r_w$ & \texttt{Linear}($d_k,K$), default initialization & Trainable \\
Update stability $W_b$ & \texttt{Linear}($d_k,K$), default initialization & Trainable \\
Visual salience $\phi$ & \texttt{Linear}($d_{\mathrm{model}},\lfloor d_{\mathrm{model}}/4\rfloor$)--GELU--\texttt{Linear}($\lfloor d_{\mathrm{model}}/4\rfloor,1$), default initialization & Trainable \\
Read conditioning $W_c$ & $d_{\mathrm{model}}\!\times\!d_{\mathrm{model}}$, bias-free; default initialization & Trainable \\
Read projection $W_r$ & $d_k\!\times\!d_{\mathrm{model}}$, copied from the initialized $W_k$ at construction & Trainable \\
Query--value alignment $W_{qv}$ & $d_v\!\times\!d_k$, bias-free; default initialization & Trainable \\
Read router $R$ & \texttt{Linear}($3d_v,d_v$)--GELU--\texttt{Linear}($d_v,1$); $3d_v^2+2d_v+1$ parameters (49,409 at $d_v{=}128$), default initialization & Trainable \\
Steering projection heads $H_{g,m}^{k, v}$ & $10^{-4}\mathcal{N}(0,1)$ & Trainable \\
Base steering scale $\hat\alpha$ & $\log\alpha_{\mathrm{train}}$ & Trainable \\
Prediction branch $(P_q,P_p,F_{\mathrm{pred}})$ & $d_vd_{\mathrm{model}}+4d_v^2+7d_v$; $P_q$ near-identity $+0.01\,\mathcal{N}(0,1)$, $P_p/F_{\mathrm{pred}}$ default Linear/LayerNorm init; 328,576 for Qwen2.5-VL-3B & Learned only when $\lambda_{\mathrm{pred}}>0$ \\
\addlinespace[3pt]
\multicolumn{3}{@{}l}{\textbf{Fixed Hyperparameters (Frozen Architectural Constants)}} \\
Base write step size $\eta_w$ & $1.0$ & Frozen constant \\
Confidence momentum $\gamma_c$ & $0.95$ & Frozen constant \\
Read-router temperature $\tau$ & $1.0$ & Frozen constant \\
\bottomrule
\end{tabularx}
\end{table}

\subsection{Feature Prediction Head}
\label{sec:appendix-prediction}

For prediction, we select a contiguous $\min(T_{\mathrm{pred}},T_w-1)$-step span with the highest detached mean salience, block its writes, and run a second writer pass to obtain a held-out state $S^{-}$. Let $T_w$ be the pooled writer sequence length, $\tilde x_t=\operatorname{LN}(x_t)\in\mathbb{R}^{d_{\mathrm{model}}}$ denote the video-only writer input, $q\in\mathbb{R}^{d_{\mathrm{model}}}$ the mean non-visual prefix embedding, and $\bar t=(t-1)/(T_w-1)\in[0,1]$ the normalized temporal position. The prediction read reuses $W_r$, $A_m$, $W_{qv}$, and $R$ from Eqs.~\ref{eq:read}--\ref{eq:read-router}, with $W_r\operatorname{LN}(q)$ as its query. The prediction branch reads the held-out state at every writer step and computes
\begin{equation}
  \begin{aligned}
    r_t^{-}&=\operatorname{Read}(S^{-},q\mathbf{1}_{T_w})_t\in\mathbb{R}^{d_v},\\
    h_t&=r_t^{-}+P_q\operatorname{LN}(q)+P_p\bar t+b_p,\\
    \hat y_t&=F_{\mathrm{pred}}(h_t),
  \end{aligned}
  \label{eq:prediction-head}
\end{equation}
Here $P_q\in\mathbb{R}^{d_v\times d_{\mathrm{model}}}$ is bias-free, $P_p\in\mathbb{R}^{d_v\times 1}$ has bias $b_p\in\mathbb{R}^{d_v}$, and $F_{\mathrm{pred}}$ is $\operatorname{LN}_{d_v}\!\to\!\operatorname{Linear}(d_v,2d_v)\!\to\!\operatorname{GELU}\!\to\!\operatorname{Linear}(2d_v,d_v)$.
The target is the stop-gradient, normalized value projection $y_t=\operatorname{sg}(\operatorname{norm}(W_v\tilde x_t))$, and the masked objective is
\begin{equation}
  \mathcal L_{\mathrm{pred}}=\frac{1}{\sum_t m_t}\sum_{t=1}^{T_w}m_t\lVert\operatorname{norm}(\hat y_t)-y_t\rVert_2^2,
  \label{eq:prediction-loss}
\end{equation}
where $m_t$ is the selected span mask. The loss is evaluated per video and averaged across the batch; sequences with $T_w\leq1$ contribute zero. Thus the head has input and output shape $[N,T_w,d_v]$ for batch size $N$; its only temporal conditioning is the scalar normalized position $\bar t$, while query conditioning is broadcast across the $T_w$ writer steps. The branch contains $d_vd_{\mathrm{model}}+4d_v^2+7d_v$ trainable parameters (328,576 for Qwen2.5-VL-3B with $d_{\mathrm{model}}{=}2048,d_v{=}128$). It is used only during training and is disabled at test time.

\subsection{Evaluation Protocols and Benchmark Specifications}
\label{sec:appendix-eval}

Evaluation is conducted at 1\,FPS with a maximum resolution cap of 200{,}704 pixels per frame. For videos up to 240 seconds, frames are sampled at 1\,FPS; for longer videos, the stream is uniformly sampled across the full temporal horizon up to the maximum ingestion cap of $T{=}240$ frames. Both the writer and the decoder share the same frame sample under identical visual resolution caps. Testing evaluates decoder visual budgets $B \in \{16, 32, 64, 90\}$; same-backbone comparisons match $B$ identically. In offline mode, the visual prompt buffer $\mathcal{V}_B$ samples $B$ frames uniformly across the stream, and the writer processes the sequence in a single pass. In streaming end-of-stream mode, frames arrive chronologically in sequential two-frame chunks carrying state $(S,c)$ forward while a deterministic uniform $B$-frame reservoir is maintained; the reservoir contents form the prompt buffer when the stream ends. All benchmarks evaluate via one-token constrained decoding over option letters $\{A, B, C, D, \dots\}$ using official evaluation scripts across 12{,}261 total questions: LongVideoBench (1{,}337), MLVU (2{,}175), Video-MME (2{,}700), EgoSchema (500), MVBench (4{,}000), and LVBench (1{,}549).

\section{Complete Experimental Results and Analyses}
\label{sec:appendix-results}

\begin{figure}[H]
  \centering
  \includegraphics[width=\textwidth]{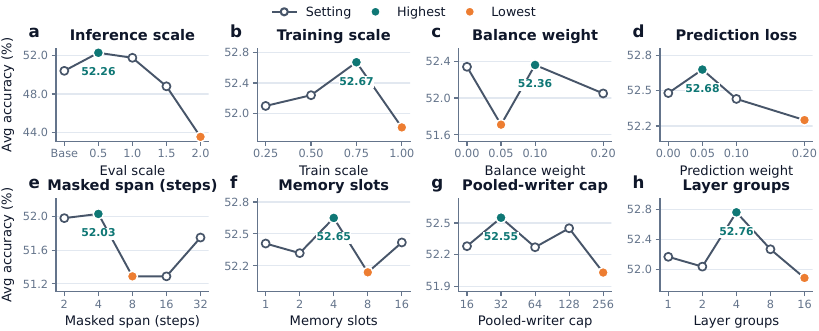}
  \caption{One-factor sensitivity sweeps on Qwen2.5-VL-3B (offline $B{=}16$, Avg). Markers denote checkpoints or inference configurations; teal and orange highlight extreme values within each sweep. Corresponding per-benchmark numeric breakdowns are tabulated in Table~\ref{tab:sen}.}
  \label{fig:sensitivity-app}
\end{figure}

This section provides the exhaustive tabular results and fine-grained breakdowns supporting the analyses in Section~\ref{sec:experiments}. Tables~\ref{tab:budget} and~\ref{tab:budget-llava} present the complete per-benchmark results across visual budgets $B \in \{16, 32, 64, 90\}$ for Qwen and LLaVA-Video backbones (supporting Figure~\ref{fig:budget-gains}). Table~\ref{tab:modulation} documents the individual benchmark accuracies across steering targets (supporting Figure~\ref{fig:ablation}a). Table~\ref{tab:writer-t} reports per-benchmark accuracies across writer training horizons. Table~\ref{tab:efficiency} reports streaming end-of-stream latency and memory. Furthermore, Figure~\ref{fig:sensitivity-app} and Table~\ref{tab:sen} document exhaustive sensitivity sweeps across 36 parameter configurations and a clean Base reference.

\subsection{Diagnostic Configurations}
\label{sec:appendix-diagnostics}

The duration analysis in Figure~\ref{fig:duration-gains} uses the $T_{\mathrm{mem}}{=}16$ checkpoint from Table~\ref{tab:sen}: $K{=}4$, $G{=}1$, training frame cap $64$, $\alpha_{\mathrm{train}}{=}\alpha_{\mathrm{run}}{=}1$, $\lambda_{\mathrm{bal}}{=}0.05$, $\lambda_{\mathrm{pred}}{=}0.1$, and $T_{\mathrm{pred}}{=}8$. Its aggregate accuracies are 54.56\% on Video-MME and 38.09\% on LVBench. The diagnostic compares the uniform-frame Base with a reservoir-buffer PReM run at $B{=}16$ and ingestion cap $240$. The component ablations in Table~\ref{tab:modulation} share the Full \ours{} setup except for the stated component change. Table~\ref{tab:sen} varies its stated factor within each sweep.

Table~\ref{tab:efficiency} reports runtime under $T{=}240$ evaluation ingestion from separate profiling runs; accuracies follow Table~\ref{tab:budget}. Latency columns are rounded independently.

\begin{figure}[H]
  \centering
  \includegraphics[width=\textwidth]{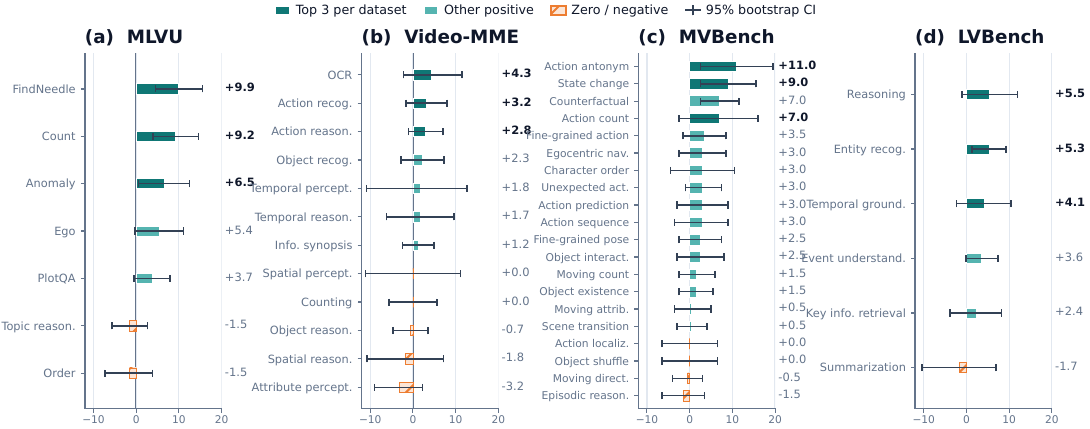}
  \caption{Exhaustive native task-category accuracy differences ($\ours{} - \mathrm{Base}$, \%) on Qwen2.5-VL-3B at offline $B{=}16$. Error bars denote 95\% paired bootstrap confidence intervals across question resamples. Benchmarks lacking native task categories (LongVideoBench, EgoSchema) are omitted.}
  \label{fig:advantage-cases-full}
\end{figure}

\begin{table}[H]
\caption{\tabtitle{Matched-budget results on Qwen backbones.} Accuracy (\%). Each cell shows Base\,$\to$\,\ours{} performance across budgets $B$. Bold highlights the superior score between paired settings; $\Delta$ denotes absolute Avg gain. $B{=}240$ is a high-budget Base reference.}
\label{tab:budget}
\centering
\scriptsize
\setlength{\tabcolsep}{3.0pt}
\renewcommand{\arraystretch}{1.08}
\resizebox{\linewidth}{!}{%
\begin{tabular}{@{}lcccccc@{\hspace{8pt}}c@{\hspace{6pt}}c@{}}
\toprule
\tabhead{Budget} & \tabhead{LVB} & \tabhead{MLVU} & \tabhead{VMME} & \tabhead{EgoS} & \tabhead{MVB} & \tabhead{LVBN} & \tabhead{Avg} & \tabhead{$\Delta$} \\
\midrule
\multicolumn{9}{@{}l}{\textbf{(a) Qwen2.5-VL-3B (Offline Ingestion)}} \\
$B{=}16$ & 49.06\,$\to$\,\textbf{53.48} & 54.90\,$\to$\,\textbf{59.82} & 53.11\,$\to$\,\textbf{54.19} & 54.40\,$\to$\,\textbf{54.60} & 56.50\,$\to$\,\textbf{60.13} & 34.34\,$\to$\,\textbf{38.48} & 50.39\,$\to$\,\textbf{53.45} & \gain{3.06} \\
$B{=}32$ & 50.86\,$\to$\,\textbf{53.33} & 58.53\,$\to$\,\textbf{62.48} & 57.26\,$\to$\,\textbf{57.44} & 56.80\,$\to$\,\textbf{58.60} & 57.77\,$\to$\,\textbf{61.50} & 37.70\,$\to$\,\textbf{40.03} & 53.15\,$\to$\,\textbf{55.56} & \gain{2.41} \\
$B{=}64$ & 51.68\,$\to$\,\textbf{55.65} & 61.47\,$\to$\,\textbf{64.23} & 59.56\,$\to$\,\textbf{59.81} & 60.60\,$\to$\,\textbf{61.40} & 57.88\,$\to$\,\textbf{61.30} & 38.99\,$\to$\,\textbf{41.06} & 55.03\,$\to$\,\textbf{57.24} & \gain{2.21} \\
$B{=}90$ & 52.13\,$\to$\,\textbf{55.27} & 63.45\,$\to$\,\textbf{65.47} & 61.26\,$\to$\,\textbf{61.30} & 61.20\,$\to$\,61.20 & 57.92\,$\to$\,\textbf{61.35} & 40.41\,$\to$\,\textbf{43.12} & 56.06\,$\to$\,\textbf{57.95} & \gain{1.89} \\
\addlinespace[2pt]
Ref ($B{=}240$) & 53.78 & 65.61 & 62.15 & 62.20 & 60.50 & 43.12 & 57.89 & -- \\
\midrule
\multicolumn{9}{@{}l}{\textbf{(b) Qwen2.5-VL-3B (Streaming End-of-Stream)}} \\
$B{=}16$ & 52.73\,$\to$\,\textbf{53.55} & 55.68\,$\to$\,\textbf{59.68} & 53.89\,$\to$\,\textbf{54.11} & 54.40\,$\to$\,\textbf{54.60} & 59.98\,$\to$\,\textbf{60.33} & 35.38\,$\to$\,\textbf{38.48} & 52.01\,$\to$\,\textbf{53.46} & \gain{1.45} \\
$B{=}32$ & 53.40\,$\to$\,\textbf{54.53} & 60.00\,$\to$\,\textbf{62.34} & 56.26\,$\to$\,\textbf{57.70} & 58.40\,$\to$\,\textbf{58.60} & 61.38\,$\to$\,\textbf{61.50} & 37.90\,$\to$\,\textbf{40.09} & 54.56\,$\to$\,\textbf{55.79} & \gain{1.23} \\
$B{=}64$ & 54.90\,$\to$\,\textbf{56.02} & 61.66\,$\to$\,\textbf{64.18} & 58.93\,$\to$\,\textbf{59.70} & 60.60\,$\to$\,\textbf{60.80} & 61.08\,$\to$\,\textbf{61.33} & 41.12\,$\to$\,\textbf{41.45} & 56.38\,$\to$\,\textbf{57.25} & \gain{0.87} \\
$B{=}90$ & 54.75\,$\to$\,\textbf{56.54} & 63.59\,$\to$\,\textbf{65.52} & 59.63\,$\to$\,\textbf{61.30} & \textbf{61.40}\,$\to$\,61.20 & 61.20\,$\to$\,\textbf{61.40} & 41.77\,$\to$\,\textbf{42.93} & 57.06\,$\to$\,\textbf{58.15} & \gain{1.09} \\
\midrule
\multicolumn{9}{@{}l}{\textbf{(c) Qwen3-VL-8B (Offline Ingestion)}} \\
$B{=}16$ & 56.84\,$\to$\,\textbf{57.52} & 60.55\,$\to$\,\textbf{62.30} & 60.00\,$\to$\,\textbf{60.41} & 69.00\,$\to$\,\textbf{70.00} & 64.60\,$\to$\,\textbf{66.15} & 38.00\,$\to$\,\textbf{38.28} & 58.16\,$\to$\,\textbf{59.11} & \gain{0.95} \\
$B{=}32$ & 58.94\,$\to$\,\textbf{59.69} & 65.15\,$\to$\,\textbf{65.29} & 63.33\,$\to$\,\textbf{64.04} & 70.00\,$\to$\,\textbf{73.00} & 65.38\,$\to$\,\textbf{67.33} & 39.83\,$\to$\,\textbf{40.67} & 60.44\,$\to$\,\textbf{61.67} & \gain{1.23} \\
$B{=}64$ & 60.36\,$\to$\,\textbf{61.85} & 68.18\,$\to$\,\textbf{70.30} & 66.11\,$\to$\,\textbf{66.96} & 70.60\,$\to$\,\textbf{74.60} & 65.83\,$\to$\,\textbf{67.03} & 41.83\,$\to$\,\textbf{44.16} & 62.15\,$\to$\,\textbf{64.15} & \gain{2.00} \\
$B{=}90$ & 60.88\,$\to$\,\textbf{63.05} & 70.94\,$\to$\,\textbf{71.22} & 67.56\,$\to$\,\textbf{67.87} & 72.60\,$\to$\,\textbf{74.20} & 65.90\,$\to$\,\textbf{67.10} & 45.32\,$\to$\,\textbf{45.76} & 63.87\,$\to$\,\textbf{64.87} & \gain{1.00} \\
\addlinespace[2pt]
Ref ($B{=}240$) & 64.25 & 75.59 & 70.19 & 73.00 & 65.95 & 47.39 & 66.06 & -- \\
\midrule
\multicolumn{9}{@{}l}{\textbf{(d) Qwen3-VL-8B (Streaming End-of-Stream)}} \\
$B{=}16$ & 56.92\,$\to$\,\textbf{57.74} & 59.17\,$\to$\,\textbf{62.53} & 58.41\,$\to$\,\textbf{60.52} & 69.00\,$\to$\,\textbf{69.40} & 64.55\,$\to$\,\textbf{66.13} & 36.99\,$\to$\,\textbf{38.22} & 57.51\,$\to$\,\textbf{59.09} & \gain{1.58} \\
$B{=}32$ & 58.56\,$\to$\,\textbf{60.58} & 62.90\,$\to$\,\textbf{65.29} & 62.52\,$\to$\,\textbf{64.00} & 71.40\,$\to$\,\textbf{72.40} & 65.83\,$\to$\,\textbf{67.25} & 40.48\,$\to$\,\textbf{40.67} & 60.28\,$\to$\,\textbf{61.70} & \gain{1.42} \\
$B{=}64$ & \textbf{62.00}\,$\to$\,61.78 & 67.26\,$\to$\,\textbf{70.34} & 65.67\,$\to$\,\textbf{66.96} & 72.20\,$\to$\,\textbf{74.60} & 66.03\,$\to$\,\textbf{67.00} & 42.87\,$\to$\,\textbf{43.71} & 62.67\,$\to$\,\textbf{64.07} & \gain{1.40} \\
$B{=}90$ & 61.71\,$\to$\,\textbf{62.15} & 69.75\,$\to$\,\textbf{71.22} & 67.22\,$\to$\,\textbf{67.87} & 72.40\,$\to$\,\textbf{74.20} & 66.05\,$\to$\,\textbf{67.13} & 42.61\,$\to$\,\textbf{45.76} & 63.29\,$\to$\,\textbf{64.72} & \gain{1.43} \\
\bottomrule
\end{tabular}}
\end{table}

\begin{table}[H]
\caption{\tabtitle{Cross-family transfer to LLaVA-Video-7B.} Accuracy (\%). Each cell shows Base\,$\to$\,\ours{} performance across visual budgets $B$. Bold highlights the superior score between paired settings; $\Delta$ denotes absolute Avg gain over Base. Evaluated through budget $B{=}90$.}
\label{tab:budget-llava}
\centering
\scriptsize
\setlength{\tabcolsep}{3.0pt}
\renewcommand{\arraystretch}{1.08}
\resizebox{\linewidth}{!}{%
\begin{tabular}{@{}lcccccc@{\hspace{8pt}}c@{\hspace{6pt}}c@{}}
\toprule
\tabhead{Budget} & \tabhead{LVB} & \tabhead{MLVU} & \tabhead{VMME} & \tabhead{EgoS} & \tabhead{MVB} & \tabhead{LVBN} & \tabhead{Avg} & \tabhead{$\Delta$} \\
\midrule
\multicolumn{9}{@{}l}{\textbf{(a) LLaVA-Video-7B (Offline Ingestion)}} \\
$B{=}16$ & 56.32\,$\to$\,\textbf{57.52} & 61.29\,$\to$\,\textbf{62.76} & 59.96\,$\to$\,\textbf{60.07} & 54.20\,$\to$\,\textbf{56.00} & 58.70\,$\to$\,\textbf{59.03} & 38.35\,$\to$\,\textbf{38.80} & 54.80\,$\to$\,\textbf{55.70} & \gain{0.90} \\
$B{=}32$ & 58.12\,$\to$\,\textbf{58.64} & 66.57\,$\to$\,\textbf{66.76} & 62.15\,$\to$\,\textbf{62.37} & 58.00\,$\to$\,\textbf{58.20} & 59.90\,$\to$\,\textbf{60.48} & 40.87\,$\to$\,\textbf{42.29} & 57.60\,$\to$\,\textbf{58.12} & \gain{0.52} \\
$B{=}64$ & 59.39\,$\to$\,\textbf{59.99} & 70.02\,$\to$\,\textbf{70.39} & 63.33\,$\to$\,\textbf{63.48} & \textbf{58.20}\,$\to$\,58.00 & 59.60\,$\to$\,\textbf{60.15} & 42.41\,$\to$\,\textbf{43.58} & 58.82\,$\to$\,\textbf{59.27} & \gain{0.45} \\
$B{=}90$ & 59.69\,$\to$\,\textbf{60.66} & 70.39\,$\to$\,\textbf{70.57} & \textbf{63.74}\,$\to$\,63.52 & 58.40\,$\to$\,\textbf{59.00} & 60.12\,$\to$\,\textbf{60.18} & 45.13\,$\to$\,\textbf{45.47} & 59.58\,$\to$\,\textbf{59.90} & \gain{0.32} \\
\midrule
\multicolumn{9}{@{}l}{\textbf{(b) LLaVA-Video-7B (Streaming End-of-Stream)}} \\
$B{=}16$ & 55.87\,$\to$\,\textbf{56.47} & 60.46\,$\to$\,\textbf{61.61} & 59.41\,$\to$\,\textbf{60.07} & 53.80\,$\to$\,\textbf{54.00} & 57.43\,$\to$\,\textbf{58.83} & 39.57\,$\to$\,\textbf{40.28} & 54.42\,$\to$\,\textbf{55.21} & \gain{0.79} \\
$B{=}32$ & 57.29\,$\to$\,\textbf{58.12} & 65.75\,$\to$\,\textbf{66.11} & \textbf{61.00}\,$\to$\,60.96 & \textbf{57.60}\,$\to$\,57.00 & 58.03\,$\to$\,\textbf{58.25} & 40.48\,$\to$\,\textbf{40.87} & 56.69\,$\to$\,\textbf{56.89} & \gain{0.20} \\
$B{=}64$ & 57.29\,$\to$\,\textbf{58.79} & 68.18\,$\to$\,\textbf{69.66} & 62.59\,$\to$\,\textbf{63.00} & 58.20\,$\to$\,\textbf{58.40} & 58.03\,$\to$\,\textbf{58.18} & 41.83\,$\to$\,\textbf{43.58} & 57.69\,$\to$\,\textbf{58.60} & \gain{0.91} \\
$B{=}90$ & 58.04\,$\to$\,\textbf{59.61} & 68.55\,$\to$\,\textbf{70.57} & 63.85\,$\to$\,\textbf{63.98} & 59.00\,$\to$\,\textbf{59.60} & 58.05\,$\to$\,\textbf{58.90} & 42.16\,$\to$\,\textbf{44.74} & 58.28\,$\to$\,\textbf{59.57} & \gain{1.29} \\
\bottomrule
\end{tabular}}
\end{table}

\begin{table}[H]
\caption{\tabtitle{Streaming end-of-stream efficiency on Qwen2.5-VL-3B.} Evaluated on NVIDIA H20 (1\,FPS, batch size 1), supporting Figure~\ref{fig:efficiency-horizon}c--d. Base and \ours{} share budget $B$. Ingestion and answering are averaged over all QA pairs, including cache hits with negligible ingestion time; total latency reports mean / p95. Subscripts indicate peak memory overhead; $\Delta$ denotes absolute Avg gain.}
\label{tab:efficiency}
\centering
\scriptsize
\setlength{\tabcolsep}{3.2pt}
\renewcommand{\arraystretch}{0.90}
\resizebox{\linewidth}{!}{%
\begin{tabular}{@{}lccccccc@{}}
\toprule
\tabhead{Method} & \tabhead{$B$} & \tabhead{\makecell{Backbone / Train.\\Params (M)}} &
\tabhead{Peak Mem.\ (GiB)} & \tabhead{Ingest / Ans.\ (s)} &
\tabhead{Mean / p95 (s)} & \tabhead{Online Avg} & \tabhead{$\Delta$} \\
\midrule
Base & 16 & 3754.62 / 0.00 & 7.43 & 5.64 / 0.30 & 5.94 / 29.78 & 52.01 & -- \\
\rowcolor{asmlight}\ours{} & 16 & 3754.62 / 9.07 & 7.46$_{\scriptscriptstyle +0.03}$ & 7.81 / 0.38 & 8.19 / 38.11 & \best{53.46} & \gain{+1.45} \\
\midrule
Base & 32 & 3754.62 / 0.00 & 7.83 & 5.71 / 0.53 & 6.24 / 30.31 & 54.56 & -- \\
\rowcolor{asmlight}\ours{} & 32 & 3754.62 / 9.07 & 7.86$_{\scriptscriptstyle +0.03}$ & 7.83 / 0.64 & 8.48 / 38.76 & \best{55.79} & \gain{+1.23} \\
\midrule
Base & 64 & 3754.62 / 0.00 & 8.64 & 5.81 / 0.96 & 6.77 / 31.44 & 56.38 & -- \\
\rowcolor{asmlight}\ours{} & 64 & 3754.62 / 9.07 & 8.67$_{\scriptscriptstyle +0.03}$ & 7.88 / 1.15 & 9.02 / 40.02 & \best{57.25} & \gain{+0.87} \\
\midrule
Base & 90 & 3754.62 / 0.00 & 9.29 & 5.89 / 1.31 & 7.19 / 32.48 & 57.06 & -- \\
\rowcolor{asmlight}\ours{} & 90 & 3754.62 / 9.07 & 9.32$_{\scriptscriptstyle +0.03}$ & 7.90 / 1.55 & 9.45 / 40.95 & \best{58.15} & \gain{+1.09} \\
\bottomrule
\end{tabular}}
\end{table}

\begin{table}[H]
\caption{\tabtitle{Diagnostic component ablations on Qwen2.5-VL-3B} (offline $B{=}16$, \%). Steering targets evaluate separately trained heads. All variants share the same experimental setup except for the ablated component. $\Delta$ denotes the score drop relative to Full \ours{}. Zero-buffer probes ($B{=}0$) isolate recurrent retention without visual prompts.}
\label{tab:modulation}
\centering
\scriptsize
\setlength{\tabcolsep}{3.2pt}
\renewcommand{\arraystretch}{1.02}
\resizebox{\linewidth}{!}{%
\begin{tabular}{@{}lcccccc@{\hspace{6pt}}c@{\hspace{6pt}}c@{}}
\toprule
\tabhead{Variant} & \tabhead{LVB} & \tabhead{MLVU} & \tabhead{VMME} &
\tabhead{EgoS} & \tabhead{MVB} & \tabhead{LVBN} & \tabhead{Avg} &
\tabhead{$\Delta$ vs. Full} \\
\midrule
Base ($B{=}16$, no memory) & 49.06 & 54.90 & 53.11 & 54.40 & 56.50 & 34.34 & 50.39 & -- \\
\rowcolor{gray!7}Full \ours{} (K/V modulation) & \textbf{53.48} & 59.82 & \textbf{54.19} & \textbf{54.60} & \textbf{60.13} & 38.48 & \textbf{53.45} & -- \\
\midrule
\multicolumn{9}{@{}l}{\textbf{Modulation Interface}} \\
\quad Q/O modulation & 52.66 & 59.49 & 53.81 & 47.00 & 59.98 & 36.02 & 51.49 & \dropval{1.96} \\
\quad Key-only (K) modulation & 52.28 & 56.14 & 53.33 & 52.00 & 59.78 & 36.41 & 51.66 & \dropval{1.79} \\
\quad Value-only (V) modulation & 51.91 & 59.86 & 53.59 & 49.80 & 59.12 & 37.25 & 51.92 & \dropval{1.53} \\
\addlinespace[3pt]
\multicolumn{9}{@{}l}{\textbf{Write Gating Terms}} \\
\quad Uniform write routing ($\pi^w$) & 52.73 & 59.36 & 53.96 & 52.40 & 59.58 & 37.96 & 52.66 & \dropval{0.79} \\
\quad w/o Update stability ($b$) & 52.06 & 57.70 & 53.48 & 53.00 & 59.60 & 37.70 & 52.26 & \dropval{1.19} \\
\quad w/o Visual salience ($z$) & 51.91 & 58.53 & 54.00 & 50.80 & 59.73 & 37.83 & 52.13 & \dropval{1.32} \\
\quad w/o Confidence protection ($a$) & 52.43 & \textbf{60.23} & 53.70 & 49.40 & 59.45 & 37.44 & 52.11 & \dropval{1.34} \\
\quad w/o Residual novelty ($n$) & 51.98 & 59.17 & 53.30 & 49.60 & 59.38 & \textbf{38.73} & 52.03 & \dropval{1.42} \\
\addlinespace[3pt]
\multicolumn{9}{@{}l}{\textbf{Zero-Buffer Retention Probes ($B{=}0$)}} \\
\quad \ours{} ($B{=}0$, memory only) & 43.40 & 47.73 & 40.65 & 27.80 & 40.88 & 29.50 & 38.33 & -- \\
\quad Base ($B{=}0$, vision-free) & 39.27 & 42.02 & 37.30 & 26.80 & 36.70 & 26.86 & 34.82 & -- \\
\bottomrule
\end{tabular}}
\end{table}

\begin{table}[H]
\caption{\tabtitle{Writer training budget.} Per-benchmark accuracy (\%) on Qwen2.5-VL-3B ($B{=}16$, evaluated at $T{=}240$), supporting Figure~\ref{fig:efficiency-horizon}a--b. Checkpoints use write cap $T$ ($\alpha_{\mathrm{run}}{=}0.75$); clean Base rows provide the protocol-matched reference. Shading and bold highlight the highest Avg per protocol.}
\label{tab:writer-t}
\centering
\scriptsize
\setlength{\tabcolsep}{3.5pt}
\renewcommand{\arraystretch}{0.90}
\resizebox{\linewidth}{!}{%
\begin{tabular}{@{}lcccccccc@{}}
\toprule
\tabhead{Training Horizon} & \tabhead{LVB} & \tabhead{MLVU} & \tabhead{VMME} & \tabhead{EgoS} & \tabhead{MVB} & \tabhead{LVBN} & \tabhead{Avg} & \tabhead{$\Delta$} \\
\midrule
\multicolumn{9}{@{}l}{\textbf{Offline Ingestion}} \\
Clean Base & 49.06 & 54.90 & 53.11 & 54.40 & 56.50 & 34.34 & 50.39 & -- \\
Writer $T{=}16$  & 52.58 & 58.25 & 53.85 & 52.80 & 59.92 & 37.96 & 52.56 & \gain{+2.17} \\
Writer $T{=}32$  & 52.58 & 57.75 & 53.78 & 52.20 & 59.98 & 38.22 & 52.42 & \gain{+2.03} \\
\rowcolor{asmlight}Writer $T{=}64$  & 53.48 & 59.82 & 54.19 & 54.60 & 60.13 & 38.48 & \best{53.45} & \gain{+3.06} \\
Writer $T{=}90$  & 53.18 & 59.82 & 54.04 & 49.80 & 60.12 & 37.44 & 52.40 & \gain{+2.01} \\
Writer $T{=}240$ & 52.36 & 58.99 & 54.07 & 52.40 & 59.52 & 38.48 & 52.64 & \gain{+2.25} \\
\midrule
\multicolumn{9}{@{}l}{\textbf{Streaming End-of-Stream}} \\
Clean Base & 52.73 & 55.68 & 53.89 & 54.40 & 59.98 & 35.38 & 52.01 & -- \\
Writer $T{=}16$  & 51.91 & 58.11 & 53.85 & 52.60 & 60.15 & 37.96 & 52.43 & \gain{+0.42} \\
Writer $T{=}32$  & 52.36 & 58.02 & 53.52 & 52.40 & 60.33 & 38.09 & 52.45 & \gain{+0.44} \\
\rowcolor{asmlight}Writer $T{=}64$  & 53.55 & 59.68 & 54.11 & 54.60 & 60.33 & 38.48 & \best{53.46} & \gain{+1.45} \\
Writer $T{=}90$  & 53.25 & 59.68 & 53.85 & 50.40 & 60.23 & 37.38 & 52.46 & \gain{+0.45} \\
Writer $T{=}240$ & 51.38 & 59.17 & 54.00 & 51.80 & 59.70 & 37.96 & 52.34 & \gain{+0.33} \\
\bottomrule
\end{tabular}}
\end{table}

\begin{table}[H]
\caption{\tabtitle{Hyperparameter sensitivity on Qwen2.5-VL-3B} (offline $B{=}16$, \%; configurations in Appendix~\ref{sec:appendix-diagnostics}). (a)~Training sweeps: $\alpha_{\mathrm{train}}$, $\lambda_{\mathrm{bal}}$, $\lambda_{\mathrm{pred}}$, and $T_{\mathrm{pred}}$ (writer steps); read-router temperature $\tau{=}1$ is fixed. (b)~Inference scale $\alpha_{\mathrm{run}}$ changes steering without retraining; Base is the clean frozen reference. (c)~Capacity sweeps: $K$ slots, training-time pooled-writer cap $T_{\mathrm{mem}}$ (no prompt positions), and $G$ layer-head groups. Shading and bold mark the highest Avg within each sweep.}
\label{tab:sen}
\centering
\scriptsize
\setlength{\tabcolsep}{2.2pt}
\renewcommand{\arraystretch}{1.04}
\begin{minipage}[t]{0.488\textwidth}
\centering
\resizebox{\linewidth}{!}{%
\begin{tabular}{@{}llcccccc@{\hspace{4pt}}c@{}}
\toprule
\tabhead{Parameter} & \tabhead{Setting} & \tabhead{LVB} & \tabhead{MLVU} &
\tabhead{VMME} & \tabhead{EgoS} & \tabhead{MVB} &
\tabhead{LVBN} & \tabhead{Avg} \\
\midrule
\multicolumn{9}{@{}l}{\textbf{(a) Training}} \\
\tablegroup{9}{Training steering scale $\alpha_{\mathrm{train}}$}
& 0.25 & 52.06 & 56.87 & 53.30 & 53.20 & 60.00 & 37.19 & 52.10 \\
& 0.50 & 52.95 & 57.52 & 54.07 & 51.80 & 59.68 & 37.44 & 52.24 \\
\rowcolor{asmlight}& 0.75 & 52.95 & 59.54 & 54.04 & 51.40 & 59.65 & 38.41 & \best{52.67} \\
& 1.00 & 51.98 & 60.18 & 53.74 & 48.60 & 58.73 & 37.70 & 51.82 \\
\midrule
\tablegroup{9}{Load-balancing weight $\lambda_{\mathrm{bal}}$}
& 0.00 & 52.36 & 60.55 & 54.15 & 49.40 & 59.30 & 38.28 & 52.34 \\
& 0.05 & 50.93 & 59.86 & 54.07 & 48.60 & 59.40 & 37.38 & 51.71 \\
\rowcolor{asmlight}& 0.10 & 53.03 & 59.54 & 53.81 & 51.20 & 59.20 & 37.38 & \best{52.36} \\
& 0.20 & 51.68 & 59.26 & 53.85 & 50.80 & 59.40 & 37.31 & 52.05 \\
\midrule
\tablegroup{9}{Prediction loss weight $\lambda_{\mathrm{pred}}$}
& 0.00 & 53.33 & 58.21 & 54.00 & 51.80 & 59.55 & 38.02 & 52.48 \\
\rowcolor{asmlight}& 0.05 & 53.03 & 58.16 & 54.00 & 52.40 & 59.88 & 38.61 & \best{52.68} \\
& 0.10 & 52.88 & 57.20 & 53.89 & 53.00 & 59.95 & 37.64 & 52.43 \\
& 0.20 & 52.21 & 56.97 & 53.67 & 53.00 & 59.95 & 37.70 & 52.25 \\
\midrule
\tablegroup{9}{Masked prediction span $T_{\mathrm{pred}}$ (writer steps)}
& 2 & 52.80 & 59.13 & 53.26 & 49.60 & 59.63 & 37.44 & 51.98 \\
\rowcolor{asmlight}& 4 & 52.95 & 59.86 & 53.93 & 48.80 & 59.30 & 37.31 & \best{52.03} \\
& 8 & 52.13 & 59.72 & 53.04 & 46.20 & 59.65 & 36.99 & 51.29 \\
& 16 & 51.68 & 60.00 & 52.89 & 47.20 & 59.23 & 36.73 & 51.29 \\
& 32 & 51.91 & 59.68 & 52.78 & 49.40 & 59.38 & 37.38 & 51.75 \\
\bottomrule
\end{tabular}}
\end{minipage}%
\hfill
\begin{minipage}[t]{0.488\textwidth}
\centering
\resizebox{\linewidth}{!}{%
\begin{tabular}{@{}llcccccc@{\hspace{4pt}}c@{}}
\toprule
\tabhead{Parameter} & \tabhead{Setting} & \tabhead{LVB} & \tabhead{MLVU} &
\tabhead{VMME} & \tabhead{EgoS} & \tabhead{MVB} &
\tabhead{LVBN} & \tabhead{Avg} \\
\midrule
\multicolumn{9}{@{}l}{\textbf{(b) Inference}} \\
\tablegroup{9}{Inference steering scale $\alpha_{\mathrm{run}}$}
& Base & 49.06 & 54.90 & 53.11 & 54.40 & 56.50 & 34.34 & 50.39 \\
\rowcolor{asmlight}& 0.5 & 52.73 & 59.22 & 53.78 & 50.00 & 59.88 & 37.96 & \best{52.26} \\
& 1.0 & 52.51 & 60.32 & 53.07 & 47.20 & 59.63 & 37.70 & 51.74 \\
& 1.5 & 49.21 & 56.87 & 50.89 & 39.80 & 58.73 & 37.25 & 48.79 \\
& 2.0 & 44.80 & 47.26 & 46.19 & 35.40 & 55.63 & 32.09 & 43.56 \\
\midrule
\multicolumn{9}{@{}l}{\textbf{(c) Memory capacity}} \\
\tablegroup{9}{Memory slot count $K$}
& 1 & 51.76 & 58.90 & 54.11 & 52.60 & 59.33 & 37.77 & 52.41 \\
& 2 & 52.43 & 59.63 & 54.37 & 50.20 & 59.33 & 37.96 & 52.32 \\
\rowcolor{asmlight}& 4 & 52.80 & 58.67 & 54.26 & 52.20 & 59.78 & 38.22 & \best{52.65} \\
& 8 & 53.25 & 58.57 & 53.59 & 50.00 & 59.23 & 38.22 & 52.14 \\
& 16 & 52.13 & 57.89 & 53.48 & 54.20 & 59.78 & 37.06 & 52.42 \\
\midrule
\tablegroup{9}{Pooled-writer cap $T_{\mathrm{mem}}$}
& 16 & 52.73 & 59.31 & 54.56 & 49.20 & 59.78 & 38.09 & 52.28 \\
\rowcolor{asmlight}& 32 & 52.88 & 59.36 & 53.89 & 52.40 & 59.88 & 36.86 & \best{52.55} \\
& 64 & 53.03 & 59.68 & 54.19 & 49.80 & 59.85 & 37.06 & 52.27 \\
& 128 & 52.95 & 59.26 & 54.11 & 51.80 & 59.53 & 37.06 & 52.45 \\
& 256 & 52.06 & 59.40 & 53.74 & 50.40 & 59.03 & 37.57 & 52.03 \\
\midrule
\tablegroup{9}{Layer-head groups $G$}
& 1 & 53.03 & 58.99 & 54.19 & 49.20 & 59.50 & 38.09 & 52.17 \\
& 2 & 52.95 & 58.48 & 53.67 & 50.00 & 59.33 & 37.83 & 52.04 \\
\rowcolor{asmlight}& 4 & 53.78 & 59.22 & 53.96 & 52.60 & 59.68 & 37.31 & \best{52.76} \\
& 8 & 52.73 & 58.99 & 53.11 & 51.80 & 59.95 & 37.06 & 52.27 \\
& 16 & 52.13 & 58.62 & 53.00 & 51.40 & 59.90 & 36.28 & 51.89 \\
\bottomrule
\end{tabular}}
\end{minipage}
\end{table}

\subsection{Multi-seed Training Stability}
\label{sec:appendix-seeds}

To verify optimization robustness, Table~\ref{tab:seeds} reports fine-grained per-benchmark accuracy alongside aggregate statistics across three independent training runs (seeds 13, 17, 23) at offline $B{=}16$. Seed 13 reuses the Table~\ref{tab:budget} main-run checkpoint. On Qwen2.5-VL-3B, mean accuracy is $53.38\%$ with sample standard deviation (SD) $0.06\%$, a $+2.99 \pm 0.06\%$ gain over Base (50.39\%); on Qwen3-VL-8B, the mean is $59.11\%$ ($\text{SD}{=}0.03\%$), a $+0.95 \pm 0.03\%$ gain over Base (58.16\%). Every seed exceeds its paired Base Avg; Table~\ref{tab:seeds} reports the per-benchmark variation.

\begin{table}[H]
\caption{\tabtitle{Multi-seed evaluation across benchmarks} (offline $B{=}16$, \%). Accuracy and standard deviations across three independent runs (seeds 13, 17, 23) alongside deterministic Base.}
\label{tab:seeds}
\centering
\scriptsize
\setlength{\tabcolsep}{2.2pt}
\renewcommand{\arraystretch}{1.04}
\begin{minipage}[t]{0.488\textwidth}
\centering
\resizebox{\linewidth}{!}{%
\begin{tabular}{@{}lccccccc@{\hspace{5pt}}c@{}}
\toprule
\tabhead{Method} & \tabhead{Seed} & \tabhead{LVB} & \tabhead{MLVU} &
\tabhead{VMME} & \tabhead{EgoS} & \tabhead{MVB} &
\tabhead{LVBN} & \tabhead{Avg} \\
\midrule
\multicolumn{9}{@{}l}{\textbf{(a) Qwen2.5-VL-3B}} \\
Base & -- & 49.06 & 54.90 & 53.11 & 54.40 & 56.50 & 34.34 & 50.39 \\
\ours{} & 13 & 53.48 & 59.82 & 54.19 & 54.60 & 60.13 & 38.48 & 53.45 \\
\ours{} & 17 & 53.33 & 59.54 & 54.15 & 54.20 & 60.15 & 38.67 & 53.34 \\
\ours{} & 23 & 53.10 & 60.64 & 53.81 & 54.00 & 59.90 & 38.67 & 53.35 \\
\midrule
\rowcolor{gray!7}\multicolumn{2}{@{}l}{\textbf{Mean}} & 53.30 & 60.00 & 54.05 & 54.27 & 60.06 & 38.61 & \textbf{53.38} \\
\multicolumn{2}{@{}l}{\color{gray}SD ($\pm$)} & {\color{gray}0.19} & {\color{gray}0.57} & {\color{gray}0.21} & {\color{gray}0.31} & {\color{gray}0.14} & {\color{gray}0.11} & {\color{gray}0.06} \\
\bottomrule
\end{tabular}}
\end{minipage}%
\hfill
\begin{minipage}[t]{0.488\textwidth}
\centering
\resizebox{\linewidth}{!}{%
\begin{tabular}{@{}lccccccc@{\hspace{5pt}}c@{}}
\toprule
\tabhead{Method} & \tabhead{Seed} & \tabhead{LVB} & \tabhead{MLVU} &
\tabhead{VMME} & \tabhead{EgoS} & \tabhead{MVB} &
\tabhead{LVBN} & \tabhead{Avg} \\
\midrule
\multicolumn{9}{@{}l}{\textbf{(b) Qwen3-VL-8B}} \\
Base & -- & 56.84 & 60.55 & 60.00 & 69.00 & 64.60 & 38.00 & 58.16 \\
\ours{} & 13 & 57.52 & 62.30 & 60.41 & 70.00 & 66.15 & 38.28 & 59.11 \\
\ours{} & 17 & 57.67 & 62.34 & 60.44 & 69.80 & 66.25 & 38.35 & 59.14 \\
\ours{} & 23 & 57.52 & 62.21 & 60.33 & 69.80 & 66.12 & 38.54 & 59.09 \\
\midrule
\rowcolor{gray!7}\multicolumn{2}{@{}l}{\textbf{Mean}} & 57.57 & 62.28 & 60.39 & 69.87 & 66.17 & 38.39 & \textbf{59.11} \\
\multicolumn{2}{@{}l}{\color{gray}SD ($\pm$)} & {\color{gray}0.09} & {\color{gray}0.07} & {\color{gray}0.06} & {\color{gray}0.12} & {\color{gray}0.07} & {\color{gray}0.13} & {\color{gray}0.03} \\
\bottomrule
\end{tabular}}
\end{minipage}
\end{table}

\section{Mathematical Notation Reference}
\label{sec:appendix-notation}

Table~\ref{tab:notation} summarizes the mathematical symbols, dimensions, and operational definitions.

\begingroup
\small
\setlength{\tabcolsep}{3.5pt}
\renewcommand{\arraystretch}{1.0}
\setlength{\LTleft}{0pt}
\setlength{\LTright}{0pt}
\setlength{\LTcapwidth}{\linewidth}
\begin{longtable}{@{}>{\raggedright\arraybackslash}p{0.15\linewidth}@{\hspace{8pt}}>{\raggedright\arraybackslash}p{\dimexpr0.85\linewidth-8pt\relax}@{}}
\caption{\tabtitle{Mathematical notation reference for Prefix-Steered Recurrent Memory.} Summarizes the core symbols, tensor dimensions, and operational definitions of \ours{} across memory updates, prefix steering, and optimization objectives.}
\label{tab:notation}
\\
\toprule
\tabhead{Symbol} & \tabhead{Meaning and Operational Scope} \\
\midrule
\endfirsthead
\multicolumn{2}{@{}l}{\small\tablename~\thetable{} (continued)}\\
\toprule
\tabhead{Symbol} & \tabhead{Meaning and Operational Scope} \\
\midrule
\endhead
\multicolumn{2}{@{}l}{\textbf{Visual Input and Recurrent Memory State}} \\
$E(v_t)_p,\ x_t$ & Encoder patch token at position $p$ and spatially pooled visual frame feature at time $t$ \\
$S \in \mathbb{R}^{K\times d_v\times d_k}$ & Recurrent associative memory state tensor across $K$ memory slots \\
$c \in [0, 1]^{K}$ & Per-slot write-confidence vector tracking temporal update history \\
$B,\ T$ & Decoder prompt visual frame budget and total writer ingestion cap (train $T{=}64$; eval $T{=}240$) \\
\addlinespace[3pt]
\multicolumn{2}{@{}l}{\textbf{Memory Write Dynamics and Gating System}} \\
$W_k, W_v, A_m$ & Learnable key/value projection matrices and slot-specific addressing transforms \\
$\kappa_{t,m}$ & Normalized write addressing query for memory slot $m$ at time $t$ \\
$e_{t,m}$ & Associative reconstruction residual $u_t - S_m\kappa_{t,m}$ \\
$b_{t,m},\ n_{t,m}$ & Update-stability gate and residual-novelty gate activations (Eq.~\ref{eq:gate}) \\
$z_t,\ a_{t,m}$ & Visual-salience gate and confidence-protection gate activations (Eq.~\ref{eq:gate}) \\
$g_{t,m}$ & Composite write coefficient: $\pi^w_{t,m} b_{t,m} n_{t,m} z_t a_{t,m}$ \\
$\sigma(f_m),\ \eta_w,\ \gamma_c$ & Learnable slot retention probability, base write step size, and confidence decay rate \\
$\pi^w$ & Softmax write-routing distribution across associative slots \\
\addlinespace[3pt]
\multicolumn{2}{@{}l}{\textbf{Prefix Steering and Read Interface}} \\
$\xi_{i,m}^{(\ell)}$ & Layer-wise normalized read query for slot $m$ at prompt token index $i$ \\
$\rho^r,\ \tau$ & Normalized read relevance scores and read-router temperature parameter \\
$H_{g,m}^{k},\ H_{g,m}^{v}$ & Group- and slot-specific key and value prefix steering projection heads \\
$d_i^{k,(\ell)},\ d_i^{v,(\ell)}$ & Additive prefix modulation vectors injected into backbone Key and Value paths \\
$W_K^{(\ell)},\ W_V^{(\ell)}$ & Frozen language model self-attention Key and Value projection weights \\
$\hat\alpha,\ \alpha_{\mathrm{run}},\ \alpha_{\mathrm{eff}}$ & Learnable scale logit, inference-time steering multiplier, and effective steering scale \\
$G,\ T_{\mathrm{mem}}$ & Attention layer group count and training-time pooled-writer cap \\
\addlinespace[3pt]
\multicolumn{2}{@{}l}{\textbf{Training Objectives and Hyperparameters}} \\
$\mathcal{L}_{\mathrm{CE}}$ & Primary cross-entropy loss over target response tokens \\
$\mathcal{L}_{\mathrm{bal}},\ \lambda_{\mathrm{bal}}$ & Router load-balancing loss (uniformity penalty) and auxiliary loss weight \\
$\mathcal{L}_{\mathrm{pred}},\ \lambda_{\mathrm{pred}}$ & Salient feature reconstruction loss and corresponding loss weight \\
$T_{\mathrm{pred}}$ & Masked prediction span measured in pooled writer steps \\
\bottomrule
\end{longtable}
\endgroup

\section{Limitations and Future Work}
\label{sec:scope}

Three design choices define the scope of this work:
\begin{itemize}[leftmargin=*, itemsep=2pt, topsep=2pt]
  \item \textbf{Fixed Slot Granularity:} The recurrent memory uses a fixed number $K$ of associative slots with fixed hidden dimensionality. Dynamic slot allocation or hierarchical multi-scale abstractions would capture complex event boundaries in continuous streams.
  \item \textbf{Frame Sampling Rate:} The architecture samples at 1\,FPS, uniformly subsampling longer videos to $T{=}240$ frames. Spatial mean-pooling retains frame summaries while discarding intra-frame layout from the recurrent state. Dense action streams at higher native frame rates call for adaptive write filtering.
  \item \textbf{Evaluation Paradigms:} Evaluation targets offline and streaming end-of-stream multiple-choice QA; future work will extend \ours{} to open-ended dialogue and lifelong streaming agent environments.
\end{itemize}

\section{LLM Usage}
\label{sec:llm-usage}

Large language models assisted with manuscript critique, methodology and experiment feedback, language editing, LaTeX grammar checking, and text formatting. The conceptual methodology, mathematical formulation, model implementation, and experimental evaluations were conducted independently by the authors. All AI-assisted text, empirical results, derivations, and factual claims were verified against the source code, training logs, and evaluation checkpoints. The authors assume full responsibility for the contents of this paper.

\end{document}